\documentclass{article}
\PassOptionsToPackage{dvipsnames}{xcolor}
\usepackage[main,final]{neurips_2026}
\usepackage[utf8]{inputenc}
\usepackage[T1]{fontenc}
\usepackage{amssymb}
\usepackage{amsmath}
\usepackage{amsbsy}
\usepackage{amsfonts}
\usepackage[nice]{nicefrac}
\usepackage{breqn}
\usepackage{mathtools}
\usepackage{graphicx}
\usepackage{subcaption}
\usepackage{svg}
\usepackage{tikz}
\usepackage{tikz-cd}
\usepackage{pgfplots}
\usepackage{float}
\usepackage{hyperref}
\usepackage{booktabs}
\usepackage{microtype}
\usepackage{multirow}
\usepackage{booktabs}
\usepackage{adjustbox}
\usepackage{soul}
\usepackage{tcolorbox}
\tcbuselibrary{skins}
\usepackage[dvipsnames]{xcolor}
\hypersetup{hidelinks}
\allowdisplaybreaks
\pgfplotsset{compat=1.16}
\newcommand\blfootnote[1]{%
  \begingroup
  \renewcommand\thefootnote{}\footnotetext{#1}%
  \addtocounter{footnote}{0}%
  \endgroup
}
\usepackage{array}
\usepackage{calc}
\newcolumntype{C}[1]{>{\centering\arraybackslash}m{#1}}
\newlength{\compcolw}
\title{Frequency-Structured Hamiltonian Neural Network for Multi-Timescale Dynamics}
\author{%
    Yaojun Li$^{*}$ \quad Yulong Yang$^{*}$ \quad Christine Allen-Blanchette\\
    Princeton University\\
    \texttt{\{yl3425, yulong.yang, ca15\}@princeton.edu}\\
}
\begin{document}
\blfootnote{$^{*}$ Equal contribution.} 
\maketitle
%%%%%%%%%%%%%%%%%%%%%%%%%%%%%%%%%%%%%%%%%%%%%%%
\begin{abstract}
    Learning coupled fast and slow dynamics accurately remains challenging for structure-preserving neural models. Hamiltonian Neural Networks preserve a learned Hamiltonian, but their neural parameterizations are susceptible to spectral bias, which favors slowly varying components and can leave fast oscillations poorly resolved. We introduce the Frequency-Structured Hamiltonian Neural Network (FS-HNN), which organizes Hamiltonian learning across temporal frequency scales. FS-HNN learns Hamiltonian components from frequency-filtered views of observed trajectories and combines them nonlinearly into a single scalar Hamiltonian. This enables the model to represent interactions across timescales while conserving the learned energy in continuous time. For PDEs, FS-HNN represents Hamiltonian components as neural functionals and learns a spatial operator, using orthogonal projection to conserve the learned energy. Experiments on ODE and PDE benchmarks spanning nonlinear oscillators, coupled lattices, shallow water flows, and dispersive waves show that FS-HNN improves long-horizon rollout accuracy over existing baselines, with accompanying gains in physical-energy fidelity. Code is available at \url{https://github.com/CAB-Lab-Princeton/FS-HNN.git}.
\end{abstract}
%%%%%%%%%%%%%%%%%%%%%%%%%%%%%%%%%%%%%%%%%%%%%%%
\section{Introduction}

Multi-timescale dynamics arise throughout engineering and science. In nonlinear lattices, fast local oscillations coexist with slow energy transfer across modes~\citep{fermi1955studies}, and shallow water flows exhibit rapid gravity waves alongside slower advective transport~\citep{vreugdenhil2013numerical}. Predicting these systems over long horizons requires resolving both timescales and their interactions. When the timescales are widely separated, models must resolve rapid oscillations over intervals long enough to capture slow evolution. Neural models face an additional challenge from spectral bias, the tendency to learn low-frequency structure before high-frequency components~\citep{rahaman2019spectral,xu2019frequency,krishnapriyan2021characterizing}. Consequently, neural models may reproduce slowly varying behavior while misrepresenting the amplitude or phase of fast oscillations, potentially distorting energy transfer between modes and degrading long-horizon accuracy in coupled systems.

Hamiltonian Neural Networks incorporate physical structure into learned dynamics by conserving a learned Hamiltonian in continuous time~\citep{greydanus2019hamiltonian,toth2019hamiltonian}. Symplectic methods impose geometric constraints through neural architectures or numerical integration schemes~\citep{zhong2019symplectic,chen2019symplectic,jin2020sympnets}. However, neither energy conservation nor symplecticity alone addresses the learning imbalance between fast and slow dynamics. Multiscale methods offer a complementary approach. Multirate and multigrid methods organize numerical computation across scales~\citep{mandel1988algebraic,mandel1990multigrid,lee2013variable}, and neural forecasting architectures model temporal structure at multiple resolutions~\citep{wu2021autoformer,wang2024timemixer}. However, these forecasting architectures do not enforce Hamiltonian structure when combining information across temporal scales. Variational multirate methods preserve mechanical structure in slow–fast systems~\citep{ober2024variational}, but do not address learning Hamiltonian representations from data. The remaining challenge is to learn and recombine information across interacting timescales while preserving Hamiltonian structure in the resulting dynamics.

We introduce the Frequency-Structured Hamiltonian Neural Network (FS-HNN), which organizes Hamiltonian learning across temporal frequency scales. FS-HNN constructs nested frequency-filtered views of observed trajectories that expose progressively coarser temporal structure while retaining shared low-frequency information. Each view provides supervision for a Hamiltonian component, and a learned nonlinear aggregator combines these components into a single scalar Hamiltonian that can represent cross-scale interactions. For ODE systems, the aggregated Hamiltonian generates dynamics through the canonical symplectic operator, conserving the learned Hamiltonian in continuous time. For PDE systems, the operator relating Hamiltonian gradients to the dynamics depends on the equation family, boundary conditions, and discretization~\citep{bridges1997multi,bridges2001multi}. FS-HNN therefore represents Hamiltonian components as neural functionals, learns a convolutional spatial operator, and uses orthogonal projection to constrain the resulting dynamics to conserve the learned Hamiltonian.

We evaluate FS-HNN on ODE and PDE benchmarks spanning integrable, chaotic, and multiscale dynamics. FS-HNN achieves lower long-horizon rollout error than the evaluated structure-preserving and black-box baselines, with accompanying gains in physical-energy fidelity. Ablations show that distinct temporal scales and nonlinear Hamiltonian aggregation improve rollout accuracy. PDE ablations further demonstrate that frequency structure and energy preservation provide complementary improvements in prediction accuracy.

Our main contributions are threefold:
\vspace{-2mm}
\begin{itemize}
    \itemsep0em
    \item We introduce a frequency-structured Hamiltonian parameterization that uses nested frequency-filtered trajectory views to learn dynamics across temporal scales.
    \item We develop a nonlinear aggregation scheme that combines these components into a single scalar Hamiltonian, allowing the model to represent cross-scale interactions while conserving the learned Hamiltonian in continuous time.
    \item We extend the framework to PDE discretizations using neural Hamiltonian functionals, a learned spatial operator, and orthogonal projection to constrain the learned dynamics to conserve the Hamiltonian.
\end{itemize}
\vspace{-2mm}
%%%%%%%%%%%%%%%%%%%%%%%%%%%%%%%%%%%%%%%%%%%%%%%
\section{Related Work}
%%%%%%%%%%%%%%%%%%%%%%%%%%%%%%%%%%%%%%%%%%%%%%%%%%%%%%
\textbf{Conservation constraints in deep learning.}
Traditional system identification algorithms achieve accurate dynamics through prescribing model classes using prior knowledge~\citep{galioto2020bayesian,galioto2020bayesian2,brunton2016discovering,epperlein2015thermoacoustics,wu2024identifying,paredes2024output,richards2024output}. Black-box learning architectures circumvent this restriction by trading interpretability for expressivity~\cite{chen2018neural,li2020fourier,lu2021learning}. Recent methods seek to strike a balance between system identification and black-box learning by embedding physical laws as inductive biases. Hamiltonian~\citep{greydanus2019hamiltonian,toth2019hamiltonian,bertalan2019learning,allen2024hamiltonian,liu2025physically,zhong2020unsupervised} and Lagrangian~\citep{lutter2019deep,cranmer2020lagrangian,allen2020lagnetvip} based neural networks parameterize the system energy and recover dynamics through the corresponding equations of motion. However, encoding energy structure alone does not prevent numerical errors from accumulating over long rollouts~\citep{chen2019symplectic,zhong2019symplectic}. Subsequent work strengthens these inductive biases either by enforcing configuration-space structure~\citep{finzi2020generalizing,zhong2020unsupervised,mason2022learning,mason2023learning,duong2024port,horn2025generalized}, which requires prior knowledge of system geometry, or by enforcing symplecticity in the dynamics. Symplecticity is typically achieved either by rolling out trajectories with symplectic integrators~\citep{chen2019symplectic,zhong2019symplectic,han2021adaptable} or by learning a symplectic map directly~\citep{jin2020sympnets,burby2020fast,chen2021data,eidnes2024pseudo}. These refinements govern how the learned dynamics are constrained and integrated, but they do not by themselves resolve the learning imbalance between fast and slow dynamics. Among these, only Chen et al.~\cite{chen2019symplectic} explicitly engages with stiff systems, studying contact discontinuities rather than the slow–fast coupling on which FS-HNN focuses. FS-HNN addresses this imbalance by structuring the Hamiltonian across temporal frequency scales, learning a separate component from each of a set of nested frequency-filtered trajectory views.

\textbf{Multiscale modeling of dynamical systems.} To mitigate the difficulty posed by stiff systems, we draw inspiration from numerical analysis, where multi-scale and multi-resolution methods have long been used to improve performance and stability for stiff dynamics~\citep{mandel1988algebraic,mandel1990multigrid,lee2013variable}.
Deep learning has adopted analogous ideas: black-box architectures decompose time series through autocorrelation~\citep{wu2021autoformer}, seasonal structure~\citep{wang2024timemixer}, or mixture-of-experts routing across timescales~\citep{hu2025adaptive,farooq2024refreshnet,yao2025solving,wu2026f3no} to improve long-horizon forecasting.
These statistical decompositions, however, do not explicitly preserve the physical structure of the underlying dynamics.
More physically informed work has incorporated fast--slow structure into NeuralODEs for singularly perturbed systems~\citep{serino2025fast}, but focuses on dissipative fast dynamics rather than the persistent fast oscillations typical of Hamiltonian systems.
From a structure-preserving perspective, Ober-Bl\"obaum et al.~\cite{ober2024variational} show that slow--fast Lagrangian systems admit variational decompositions whose components retain Lagrangian dynamics.
Their analysis is theoretical, however, and does not address how to realize such decompositions in neural architectures.
FS-HNN addresses this gap by combining multi-timescale supervision with a structure-preserving Hamiltonian parameterization, allowing cross-scale interactions to be learned within a single conservative model.
%%%%%%%%%%%%%%%%%%%%%%%%%%%%%%%%%%%%%%%%%%%%%%%%%%%%%%

%%%%%%%%%%%%%%%%%%%%%%%%%%%%%%%%%%%%%%%%%%%%%%%
\section{Background}\label{sec:background}
%%%%%%%%%%%%%%%%%%%%%%%%%%%%%%%%%%%%%%%%%%%%%%%%%%%%%%
FS-HNN incorporates Hamiltonian energy conservation and frequency-structured learning to model coupled multi-timescale dynamics. The skew-symmetric structure of Hamilton's equations motivates the energy-conserving formulation of FS-HNN, while differences in characteristic frequencies motivate its nested frequency-filtered trajectory views.

\textbf{Symplectic structure and energy conservation.} Hamiltonian mechanics reframes classical dynamics in the language of energy~\citep{Goldstein2002ClassicalMechanics, marion2013classical}. Instead of tracking the forces acting on a system, it summarizes the entire dynamics in a single scalar function, the Hamiltonian $\mathcal{H}\left(\mathbf{q}, \mathbf{p}\right)$, which maps generalized coordinates $\mathbf{q}$ and momenta $\mathbf{p}$ to total energy. The gradients of $\mathcal{H}$ with respect to coordinates and momenta then specify how the system evolves, through Hamilton's equations,
\begin{equation}
    \frac{d\mathbf{q}}{dt}=\frac{\partial\mathcal{H}\left(\mathbf{q}, \mathbf{p}\right)}{\partial\mathbf{p}}, \quad\quad \frac{d\mathbf{p}}{dt}=-\frac{\partial\mathcal{H}\left(\mathbf{q}, \mathbf{p}\right)}{\partial\mathbf{q}}.\label{eq:hamilton_equation}
\end{equation}
Letting $\mathbf{z}=\left(\mathbf{q},\mathbf{p}\right)$ denote a point in the phase space of a dynamical system, Hamilton's equations take the symplectic form
\begin{equation}
    \dot{\mathbf{z}} = \mathbf{J}\nabla_{\mathbf{z}}\mathcal{H}\left(\mathbf{z}\right), \quad\quad \mathbf{J} = \begin{bmatrix} 0 & \mathbf{I}\\ -\mathbf{I} & 0\end{bmatrix},
\end{equation}
where $\mathbf{J}$ is the skew-symmetric canonical symplectic matrix. Energy conservation then follows by differentiating $\mathcal{H}$ along a trajectory and using the skew-symmetry of $\mathbf{J}$, which gives
\begin{equation}
    \frac{d\mathcal{H}}{dt} = \nabla_{\mathbf{z}} \mathcal{H}\left(\mathbf{z}\right)^{{\intercal}}\dot{\mathbf{z}} = \nabla_{\mathbf{z}}\mathcal{H}\left(\mathbf{z}\right)^{{\intercal}}\mathbf{J}\nabla_{\mathbf{z}}\mathcal{H}\left(\mathbf{z}\right) = 0.
\end{equation}

\textbf{Multi-timescale Hamiltonian systems.} Many Hamiltonian systems of practical interest evolve on multiple, widely separated characteristic timescales~\citep{hairer2006geometric,sanders2007averaging}. For a stiff Hamiltonian system, let $\varepsilon\ll1$ be the stiffness parameter of the fast potential, and suppose the phase space splits into slow coordinates $\left(\mathbf{q}^{\mathrm{s}}, \mathbf{p}^{\mathrm{s}}\right)$ and fast coordinates $\left(\mathbf{q}^{\mathrm{f}}, \mathbf{p}^{\mathrm{f}}\right)$. The Hamiltonian then takes the form
\begin{equation}
    \mathcal{H}\left(\mathbf{q}^{\mathrm{s}}, \mathbf{p}^{\mathrm{s}}, \mathbf{q}^{\mathrm{f}}, \mathbf{p}^{\mathrm{f}}\right) = \frac{1}{2} \left(\mathbf{p}^{\mathrm{s}}\right)^{\intercal} \mathbf{M}_{\mathrm{s}}^{-1}\mathbf{p}^{\mathrm{s}}+\tfrac12 \left(\mathbf{p}^{\mathrm{f}}\right)^{\intercal} \mathbf{M}_{\mathrm{f}}^{-1}\mathbf{p}^{\mathrm{f}} + V\left(\mathbf{q}^{\mathrm{s}},\mathbf{q}^{\mathrm{f}}\right)+\tfrac{1}{\varepsilon}\bar{W}\left(\mathbf{q}^{\mathrm{f}}\right),
\end{equation}
where the stiff potential $\nicefrac{1}{\varepsilon}\bar{W}(\mathbf{q}^{\mathrm{f}})$ drives fast oscillations and $V\left(\mathbf{q}^{\mathrm{s}},\mathbf{q}^{\mathrm{f}}\right)$ couples the two subsystems. The characteristic frequency of the fast subsystem follows from a leading-order analysis of the stiff potential. Expanding $\bar{W}$ to second order about a local minimizer $\mathbf{q}_{*}^{\mathrm{f}}$ and writing $\boldsymbol{\eta}=\mathbf{q}^{\mathrm{f}} - \mathbf{q}_{*}^{\mathrm{f}}$, the fast equation of motion reduces to
\begin{equation}
    \ddot{\boldsymbol{\eta}} \approx -\varepsilon^{-1}\mathbf{M}_{\mathrm{f}}^{-1}\nabla^2\bar{W}\left(\mathbf{q}_{*}^{\mathrm{f}}\right)\boldsymbol{\eta},\label{eq:fast_oscillator_H}
\end{equation}
so the fast modes oscillate with characteristic frequency $\omega\sim\varepsilon^{-1/2}$ while the slow modes evolve on $\mathcal{O}\left(1\right)$ timescales.

\textbf{Coupling and frequency structure.} In the idealized case of vanishing coupling (i.e., when $V$ depends only on $\mathbf{q}^{\mathrm{s}}$), the Hamiltonian separates additively into independent slow and fast components,
\begin{equation}
    \mathcal{H}\left(\mathbf{q}^{\mathrm{s}}, \mathbf{p}^{\mathrm{s}}, \mathbf{q}^{\mathrm{f}}, \mathbf{p}^{\mathrm{f}}\right) = \mathcal{H}_{\mathrm{s}}\left(\mathbf{q}^{\mathrm{s}}, \mathbf{p}^{\mathrm{s}}\right) + \mathcal{H}_{\mathrm{f}}\left(\mathbf{q}^{\mathrm{f}}, \mathbf{p}^{\mathrm{f}}\right).
\end{equation}
However, in general $V$ couples slow and fast coordinates and the additive form holds only as a leading-order approximation ~\citep{ober2024variational,gauckler2013energy,bustamante2019exact}. The timescale separation nonetheless has a direct frequency-domain consequence. Because slow and fast components evolve at characteristically different rates, multi-timescale trajectories concentrate their spectral power in well-separated regions of the temporal frequency spectrum, even when coupling prevents a clean modal decomposition. FS-HNN leverages this observation through a frequency-structured parameterization based on nested frequency-filtered trajectory views, with Hamiltonian components recombined nonlinearly to account for the coupling. Additional details are provided in Appendix~\ref{app:slowfast}.
%%%%%%%%%%%%%%%%%%%%%%%%%%%%%%%%%%%%%%%%%%%%%%%%%%%%%%
%%%%%%%%%%%%%%%%%%%%%%%%%%%%%%%%%%%%%%%%%%%%%%%
\section{Method}
%%%%%%%%%%%%%%%%%%%%%%%%%%%%%%%%%%%%%%%%%%%%%%%%%%%%%%
\begin{figure*}[t]
    \centering
    \includegraphics[width=\linewidth]{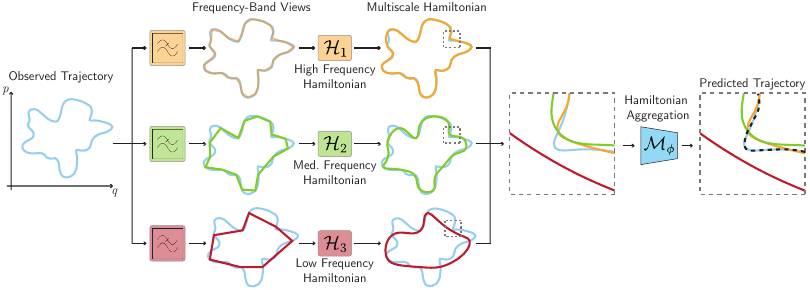}
    \caption{\textbf{Frequency-Structured Hamiltonian Neural Network.} An observed trajectory $\mathbf{z}_{\mathrm{train}}$ is filtered into a nested family of low-pass views $\mathbf{z}^{(k)}$ at decreasing cutoffs, which expose progressively coarser temporal structure. Each view provides supervision for a per-band Hamiltonian $\mathcal{H}^{(k)}_{\theta_{k}}$ that models the dynamics resolved at its cutoff. The aggregator $\mathcal{M}$ combines the per-band Hamiltonians into a single scalar Hamiltonian $\mathcal{H}_{\mathrm{agg}}$, whose canonical dynamics conserve the learned energy and generate the predicted trajectory.}
    \label{fig:MGHNN_IntroCartoon}
    \vspace{-0.15in}
\end{figure*}
The Frequency-Structured Hamiltonian Neural Network (FS-HNN) organizes Hamiltonian learning across temporal frequency scales, as illustrated in Figure~\ref{fig:MGHNN_IntroCartoon}. For ODE systems, nested frequency-filtered views of observed trajectories provide supervision for individual Hamiltonian components. A learned nonlinear aggregator combines these components into a single scalar Hamiltonian, whose canonical dynamics conserve the learned Hamiltonian in continuous time (Section~\ref{subsec:method_freq_ham}). For PDE systems, FS-HNN represents the Hamiltonian components as neural functionals, learns a problem-dependent spatial operator, and uses orthogonal projection to constrain the predicted dynamics to conserve the learned Hamiltonian (Section~\ref{subsec:method_pde}).
%%%%%%%%%%%%%%%%%%%%%%%%%%%%%%%%%%%%%%%%%%%%%%%%%%%%%%
\subsection{Frequency-Structured Parameterization of the Hamiltonian}\label{subsec:method_freq_ham}
FS-HNN incorporates the frequency structure of multi-timescale dynamics into its Hamiltonian parameterization. Rather than parameterizing the Hamiltonian with a single network, FS-HNN learns components associated with nested frequency-filtered trajectory views and combines them into a single scalar Hamiltonian. Nonlinear aggregation allows FS-HNN to represent interactions between coupled slow and fast modes.

\textbf{ODE Frequency-Filtered Views.} We construct $K$ nested low-pass-filtered views of the training trajectory $\mathbf{z}_{\mathrm{train}}$ using  decreasing cutoff frequencies $\omega_{1}>\omega_{2}>\ldots>\omega_{K}$, with $\omega_1$ equal to the Nyquist frequency:
\begin{equation}
    \mathbf{z}^{(k)} = \mathcal{F}^{-1}\left(
    \boldsymbol{1}_{\left\{\lvert\omega\rvert\leq\omega_{k}\right\}}
    \cdot\mathcal{F}\left(\mathbf{z}_{\mathrm{train}}\right)
    \right),
\end{equation}
where $\mathcal{F}$ denotes the discrete Fourier transform and $\boldsymbol{1}_{\left\{\lvert\omega\rvert\leq\omega_{k}\right\}}$ selects frequencies below $\omega_{k}$. where $\mathcal{F}$ denotes the discrete Fourier transform and $\boldsymbol{1}_{\left\{\lvert\omega\rvert\leq\omega_{k}\right\}}$ retains frequency components at or below $\omega_{k}$. The resulting views have nested spectral support: $\mathbf{z}^{(1)}$ retains the full spectrum, while each subsequent view removes additional high-frequency components. 

The default configuration uses $K=3$ nested views with cutoff frequencies of $\{1.0,0.5,0.25\}$ times the Nyquist frequency. We also evaluate spectrum-adaptive cutoffs to account for differences in spectral content across systems. For spectra with a distinct dominant peak at normalized frequency $f_{\mathrm{peak}}$, we use $\{1,f_{\mathrm{peak}}/2,f_{\mathrm{peak}}/20\}$. For broadband spectra, we use $\{1,q_{0.99},q_{0.90}\}$, where $q_{\alpha}$ denotes the normalized frequency below which a fraction $\alpha$ of the total spectral power lies. 

\textbf{Per-Band Hamiltonian.}
Each frequency-filtered trajectory is used to train a separate Hamiltonians $\mathcal{H}^{(k)}_{\theta_{k}}$ parameterized by MLPs, which learns a Hamiltonian approximation to the dynamics resolved by its filtered trajectory. Because the views are nested rather than disjoint, each branch captures the dynamics resolved at its cutoff while retaining all lower-frequency content.

\textbf{Aggregation.} A natural starting point for combining $K$ per-band Hamiltonians into a single dynamical system is the additive form
\begin{equation}
    \mathcal{H}_{\mathrm{add},\theta}\left(\mathbf{z}\right) = \sum_{k=1}^{K}\mathcal{H}^{(k)}_{\theta_{k}}\left(\mathbf{z}^{(k)}\right),
\end{equation}
which is exact when the system admits a clean slow-fast decomposition (Appendix~\ref{app:slowfast}) and the subsystems decouple. Coupling through $V$ breaks this additivity in general, so we parameterize the aggregated Hamiltonian $\mathcal{H}_{\mathrm{agg},\theta}\left(\mathbf{z}\right)$ as a learned scalar function of the per-band Hamiltonians,
\begin{equation}
    \mathcal{H}_{\mathrm{agg},\theta}\left(\mathbf{z}\right) = \mathcal{M}_{\phi}\left(\mathcal{H}^{(1)}_{\theta_{1}}\left(\mathbf{z}^{(1)}\right),\ldots,\mathcal{H}^{(K)}_{\theta_{K}}\left(\mathbf{z}^{(K)}\right)\right),
\end{equation}
where $\mathcal{M}_{\phi}$ is a small MLP. The additive case is recovered when $\mathcal{M}_{\phi}=\sum_{k}$, yielding the additive case of $\mathcal{H}_{\mathrm{agg},\theta}=\mathcal{H}_{\mathrm{add},\theta}$. Outside that limit, $\mathcal{M}_{\phi}$ captures non-additive interactions among the per-band components. The aggregated Hamiltonian then generates the dynamics through the canonical symplectic form
\begin{equation}
    \dot{\mathbf{z}} = \mathbf{J}\nabla_{\mathbf{z}}\mathcal{H}_{\mathrm{agg},\theta}\left(\mathbf{z}\right).
\end{equation}

\textbf{Structure Preservation.} Although the aggregation function $\mathcal{M}_{\phi}$ is nonlinear, $\mathcal{H}_{\mathrm{agg},\theta}$ remains a scalar Hamiltonian, and the dynamics factor through the canonical skew-symmetric $\mathbf{J}$. The conservation argument of Section~\ref{sec:background} then carries over directly, giving
\begin{equation}
    \frac{d}{dt}\mathcal{H}_{\mathrm{agg},\theta}\left(\mathbf{z}\left(t\right)\right) = \nabla_{\mathbf{z}}\mathcal{H}_{\mathrm{agg},\theta}^{\intercal}\mathbf{J}\nabla_{\mathbf{z}}\mathcal{H}_{\mathrm{agg},\theta} = 0,
\end{equation}
along trajectories. The learned aggregation therefore enriches the functional form of the Hamiltonian relative to the additive decomposition without breaking Hamiltonian structure or energy conservation in the continuous-time limit.

\textbf{Training.} FS-HNN is trained by minimizing the standard HNN loss function
\begin{equation}
    \mathcal{L} = \lVert \dot{\mathbf{z}}-\mathbf{J}\nabla_{\mathbf{z}}\mathcal{H}_{\mathrm{agg},\theta}\left(\mathbf{z}\right)\rVert^{2},
\end{equation}
on the unfiltered training trajectory. The per-band Hamiltonian $\mathcal{H}^{(k)}_{\theta_{k}}$ and aggregation function $\mathcal{M}_{\phi}$ are trained jointly through $\mathcal{L}$.
%%%%%%%%%%%%%%%%%%%%%%%%%%%%%%%%%%%%%%%%%%%%%%%%%%%%%%
\subsection{Extension to PDE Systems}\label{subsec:method_pde}
The construction in Section~\ref{subsec:method_freq_ham} leans the fact that the symplectic operator $\mathbf{J}$ is fixed, finite-dimensional, and skew-symmetric by construction. As a result $\langle\nabla\mathcal{H}, \mathbf{J}\nabla\mathcal{H}\rangle = 0$ holds automatically and the per-band Hamiltonians fully determine the dynamics. 
In the PDE setting, the state $\mathbf{z}$ is a continuous field, and in many discretizations this fixed canonical structure is no longer available. The corresponding operator becomes a differential object (e.g., $\partial_{x}$) whose form depends on the equation family, boundary conditions, and discretization scheme~\citep{bridges1997multi,bridges2001multi}, and is rarely available in closed form. We therefore learn its action and recover energy conservation through projection rather than by construction. The frequency-filtered views and the aggregation of Section~\ref{subsec:method_freq_ham} carry over unchanged, while each per-band Hamiltonian becomes a neural functional.

\textbf{Convolutional Spatial Operator.} Rather than constructing $\mathbf{J}$ analytically, we learn its action directly through a spatial operator $\mathbf{J}_{\theta}$ that maps the gradient of the aggregated Hamiltonian to the rate of change of the state,
\begin{equation}
    \Delta\tilde{\mathbf{z}} = \mathbf{J}_{\theta}\left(\nabla_{\mathbf{z}} \mathcal{H}_{\mathrm{agg},\theta}\left(\mathbf{z}\right)\right).
\end{equation}
This avoids explicit matrix representations of the operator. We implement $\mathbf{J}_{\theta}$ as a residual CNN~\citep{he2016deep}, since local convolutional kernels naturally capture neighborhood coupling on discretized spatial grids and emulate differential operators.

\textbf{Energy Preservation via Orthogonal Projection.}
Unlike the canonical $\mathbf{J}$, the learned operator $\mathbf{J}_{\theta}$ is not skew-symmetric by construction, so the conservation identity of the canonical setting no longer holds automatically. Rather than constraining $\mathbf{J}_{\theta}$ to a prescribed skew-symmetric or Poisson form, we enforce energy conservation directly by projecting its output $\Delta\tilde{\mathbf{z}}$ onto the subspace orthogonal to $\nabla_{\mathbf{z}}\mathcal{H}_{\mathrm{agg},\theta}(\mathbf{z})$,
\begin{equation}\label{eq:orth_proj}
    \Delta\mathbf{z} = \Delta\tilde{\mathbf{z}} -
    \frac{
        \langle \nabla_{\mathbf{z}}\mathcal{H}_{\mathrm{agg},\theta}(\mathbf{z}), \Delta\tilde{\mathbf{z}}\rangle
    }{
        \langle \nabla_{\mathbf{z}}\mathcal{H}_{\mathrm{agg},\theta}(\mathbf{z}),
        \nabla_{\mathbf{z}}\mathcal{H}_{\mathrm{agg},\theta}(\mathbf{z})\rangle + \epsilon
    }
    \nabla_{\mathbf{z}}\mathcal{H}_{\mathrm{agg},\theta}(\mathbf{z}),
\end{equation}
where $\Delta\mathbf{z}\approx\dot{\mathbf{z}}$ is the projected vector-field estimate and $\epsilon\ll 1$ provides numerical stability.
Writing $\mathbf{g}=\nabla_{\mathbf{z}}\mathcal{H}_{\mathrm{agg},\theta}(\mathbf{z})$, the projection satisfies
\begin{equation}
    \langle\mathbf{g},\Delta\mathbf{z}\rangle
    =
    \frac{\epsilon}{\|\mathbf{g}\|^2+\epsilon}
    \langle\mathbf{g},\Delta\tilde{\mathbf{z}}\rangle.
\end{equation}
Thus, for $\epsilon=0$ and $\mathbf{g}\neq0$, the projected vector field is exactly orthogonal to the Hamiltonian gradient, giving
\begin{equation}
    \frac{d}{dt}\mathcal{H}_{\mathrm{agg},\theta}(\mathbf{z})=\left\langle\nabla_{\mathbf{z}}\mathcal{H}_{\mathrm{agg},\theta}(\mathbf{z}),\dot{\mathbf{z}}\right\rangle=0,
\end{equation}
when the continuous-time dynamics follow the projected field.
A positive $\epsilon$ leaves the explicit conservation residual shown above.
Moreover, away from critical points, exact orthogonality implies the existence of a skew-symmetric operator $\mathbf{J}(\mathbf{z})$ satisfying
$\Delta\mathbf{z}=\mathbf{J}(\mathbf{z})\mathbf{g}$, placing the projected dynamics within the skew-gradient class associated with energy-preserving integration~\citep{mclachlan1999geometric,celledoni2012preserving}.
Thus, in the $\epsilon=0$ continuous-time limit, the construction conserves the learned Hamiltonian without requiring $\mathbf{J}_{\theta}$ itself to be skew-symmetric. It does not, however, guarantee symplecticity or exact conservation after numerical time discretization (Appendix~\ref{app:symplectic}).
Importantly, $\mathcal{H}_{\mathrm{agg},\theta}$ is a learned invariant and need not equal the physical energy. Tables~\ref{tab:ode_energy_stats} and \ref{tab:pde_energy_stats} therefore evaluate the analytic energies of the benchmark systems on the predicted states.

\textbf{PDE Frequency-Band Partitions.}
We apply temporal filtering (Section~\ref{subsec:method_freq_ham}) to PDE systems, producing nested views that retain progressively less fine-scale temporal variation.
The use of overlapping and nested views is analogous to multiscale techniques in PDE analysis, including multiresolution analysis with nested approximation spaces~\citep{mallat1989multiresolution}, multigrid methods across mesh resolutions~\citep{brandt1977multi}, and variational coupling of resolved and unresolved scales~\citep{hughes1995multiscale}.
Similarly, our construction preserves interactions across scales instead of treating frequency components as independent.

\textbf{Per-Band Hamiltonian Functional.}
Whereas each ODE Hamiltonian $\mathcal{H}^{(k)}_{\theta_k}$ is a scalar function represented by an MLP, a PDE Hamiltonian is a functional of a field. asWe therefore represent each $\mathcal{H}^{(k)}_{\theta_k}$ with a DeepONet~\citep{lu2021learning}, mapping a field representation to a scalar Hamiltonian. The resulting neural functionals are aggregated and differentiated to construct the structure-preserving PDE dynamics.
%%%%%%%%%%%%%%%%%%%%%%%%%%%%%%%%%%%%%%%%%%%%%%%%%%%%%%

%%%%%%%%%%%%%%%%%%%%%%%%%%%%%%%%%%%%%%%%%%%%%%%
\section{Experiments}
%%%%%%%%%%%%%%%%%%%%%%%%%%%%%%%%%%%%%%%%%%%%%%%%%%%%%%
\begin{figure}[b]
    \vspace{-0.15in}
    \centering
    \includegraphics[width=0.95\columnwidth]{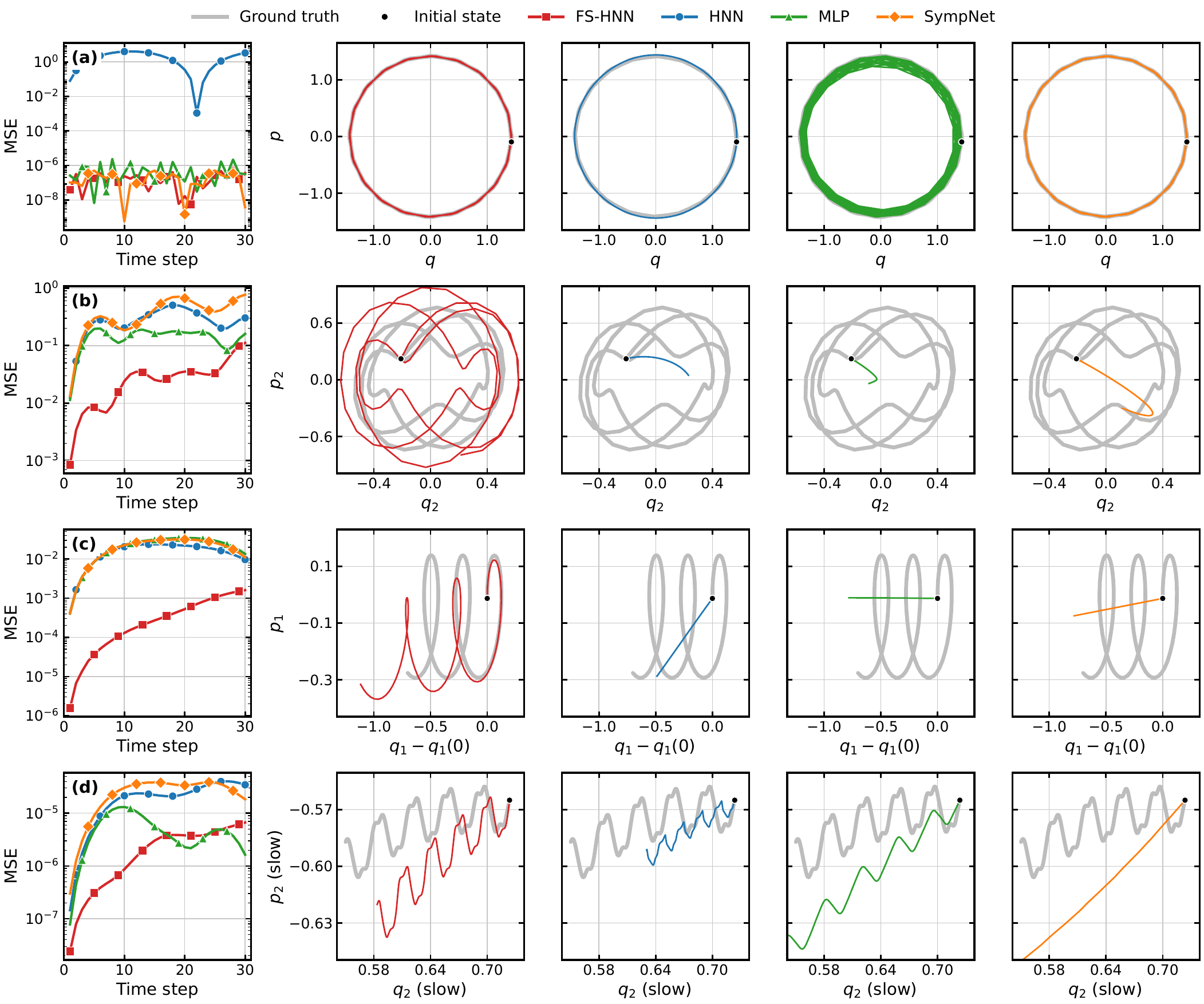}
    \vskip -0.04in
    \caption{\textbf{ODE systems rollout comparison.} Rollout MSE and phase-space portraits of FS-HNN, HNN, MLP, and SympNet on the \textbf{(a)} single pendulum, \textbf{(b)} double pendulum, \textbf{(c)} Fermi--Pasta--Ulam--Tsingou chain, and \textbf{(d)} fast--slow oscillator, shown in its slow coordinates. FS-HNN attains the lowest rollout MSE on all four systems (Table~\ref{tab:ode}).}
    \label{fig:ode}
\end{figure}
%%%%%%%%%%%%%%%%%%%%%%%%%%%%%%%%%%%%%%%%%%%%%%%%%%%%%%
The experiments examine how the frequency-structured Hamiltonian components developed above affect long-horizon rollout accuracy and physical-energy behavior in ODE and PDE systems. On ODE benchmarks ranging from integrable mechanics to chaotic and multiscale dynamics, FS-HNN is compared with black-box and structure-preserving baselines. On PDE system, the extension with its learned operator and projection is compared with structure-preserving and neural-operator baselines.
%%%%%%%%%%%%%%%%%%%%%%%%%%%%%%%%%%%%%%%%%%%%%%%%%%%%%%
\subsection{ODE Systems}
The four ODE benchmarks range from simple periodic motion to systems in which fast and slow dynamics are explicitly coupled. The single pendulum has no separation of timescales, while the double pendulum introduces chaos and sensitivity to initial conditions. The Fermi--Pasta--Ulam--Tsingou (FPUT) chain~\cite{fermi1955studies} combines fast oscillations with slow energy transfer across modes. The fast--slow oscillator proposed in~\citep{duruisseaux2023approximation} couples a unit-frequency oscillator to a slow oscillator of frequency $\varepsilon=10^{-2}$ through a nonlinear potential. The fast mode of this system lies at only $0.032$ of the Nyquist frequency, so FS-HNN uses the spectrum-selected cutoffs $\{1.0, 0.016, 0.0016\}$ of Section~\ref{subsec:method_freq_ham} on this system to align the frequency hierarchy with its spectral scales. Models are trained on the first 10 steps and rolled out for 1000 steps. We compare against a black-box MLP, HNN~\citep{greydanus2019hamiltonian}, and SympNet~\citep{jin2020sympnets}. Rollout error for FS-HNN and baselines are reported in Table~\ref{tab:ode} and Figure~\ref{fig:ode}.

\begin{table}[t]
    \centering
    \caption{\textbf{Rollout accuracy for ODE systems.} We report the rollout MSE of FS-HNN, MLP, HNN, and SympNet on the pendulum, double pendulum, Fermi--Pasta--Ulam--Tsingou, and fast--slow oscillator systems. FS-HNN achieves the lowest prediction error on all four benchmarks.}
    \label{tab:ode}
    \vskip 0.03in
    \setlength{\tabcolsep}{4pt}
    \begin{tabular}{lcccc}
        \toprule
        Model & Single Pend. & Double Pend. & FPUT & Fast--Slow \\
        \midrule
        FS-HNN & $\mathbf{(3.25{\scriptstyle\pm1.02}){\scriptstyle\times10^{-7}}}$ & $\mathbf{(3.73{\scriptstyle\pm1.50}){\scriptstyle\times10^{-2}}}$ & $\mathbf{(2.10{\scriptstyle\pm1.03}){\scriptstyle\times10^{-5}}}$ & $\mathbf{(1.45{\scriptstyle\pm0.63}){\scriptstyle\times10^{-2}}}$\\
        MLP & $(3.51{\scriptstyle\pm1.11}){\scriptstyle\times10^{-4}}$ & $(4.99{\scriptstyle\pm1.87}){\scriptstyle\times10^{-1}}$ & $(3.28{\scriptstyle\pm1.31}){\scriptstyle\times10^{-2}}$ & $(3.30{\scriptstyle\pm0.18}){\scriptstyle\times10^{-2}}$\\
        HNN & $(9.27{\scriptstyle\pm1.06}){\scriptstyle\times10^{-4}}$ & $(6.39{\scriptstyle\pm3.11}){\scriptstyle\times10^{-1}}$ & $(1.06{\scriptstyle\pm0.38}){\scriptstyle\times10^{-2}}$ & $(5.75{\scriptstyle\pm2.80}){\scriptstyle\times10^{-2}}$\\
        SympNet & $(3.26{\scriptstyle\pm1.85}){\scriptstyle\times10^{-6}}$ & $(4.56{\scriptstyle\pm1.58}){\scriptstyle\times10^{-1}}$ & $(3.43{\scriptstyle\pm0.05}){\scriptstyle\times10^{-3}}$ & $(3.48{\scriptstyle\pm0.06}){\scriptstyle\times10^{-2}}$\\
        \bottomrule
    \end{tabular}
    \vspace{-0.15in}
\end{table}

\textbf{Rollout accuracy.}
FS-HNN attains the lowest rollout MSE on all four systems (Table~\ref{tab:ode}). On the single pendulum, the phase-space portrait of FS-HNN remains stable over the full rollout while the MLP drifts visibly (Figure~\ref{fig:ode}a), so the frequency-structured parameterization retains stable rollouts even when pronounced timescale separation is absent. On the chaotic double pendulum, HNN and SympNet diverge from the reference trajectory within the first few hundred steps, whereas FS-HNN tracks it for several hundred steps before phase error accumulates (Figure~\ref{fig:ode}b). The two coupled-timescale systems test the frequency structure directly. On FPUT, FS-HNN retains the slow energy-exchange envelope that the baselines flatten or distort (Figure~\ref{fig:ode}c), and on the fast--slow oscillator it follows the slow drift of the coupled oscillator while the baselines lag behind or depart from the reference (Figure~\ref{fig:ode}d). These results support the use of frequency-structured Hamiltonian components to capture interacting fast and slow dynamics.

\textbf{Energy preservation across systems.}
The nonlinear aggregator retains the Hamiltonian structure described in Section~\ref{subsec:method_freq_ham}, so the learned Hamiltonian is conserved in continuous time. Because this learned invariant need not coincide with the physical energy, each system’s analytic energy is evaluated on the predicted states and compared with that of the reference trajectory, separately from the full-state error. FS-HNN has the lowest reported energy error across all four systems over the 1000-step rollout in Table~\ref{tab:ode_energy_stats} in Appendix~\ref{oderesults}.
%%%%%%%%%%%%%%%%%%%%%%%%%%%%%%%%%%%%%%%%%%%%%%%%%%%%%%
\begin{table}[b]
    \centering
    \caption{\textbf{Rollout accuracy for PDE systems.} We report the rollout MSE of FS-HNN, ENO, PHNN, and FNO on SWE with Gaussian-pulse and random initial conditions and on the KdV equation. The KdV entries come from single training runs on a shared train/validation split and are averaged over 20 held-out trajectories. FS-HNN achieves the lowest prediction error on all three benchmarks.}
    \vskip 0.03in
    \begin{tabular}{lccc}
        \toprule
        Model & SWE Pulse & SWE Random & KdV \\
        \midrule
        FS-HNN & $\mathbf{(3.67{\scriptstyle\pm0.54}){\scriptstyle\times10^{-4}}}$ & $\mathbf{(2.86{\scriptstyle\pm0.13}){\scriptstyle\times10^{-5}}}$ & $\mathbf{(4.62{\scriptstyle\pm4.05}){\scriptstyle\times10^{-2}}}$\\
        ENO & $(4.41{\scriptstyle\pm2.68}){\scriptstyle\times10^{-3}}$ & $(6.88{\scriptstyle\pm2.27}){\scriptstyle\times10^{-2}}$ & $(7.68{\scriptstyle\pm3.89}){\scriptstyle\times10^{-1}}$ \\
        PHNN & $(8.57{\scriptstyle\pm3.91}){\scriptstyle\times10^{-2}}$ & $(4.65{\scriptstyle\pm2.16}){\scriptstyle\times10^{-2}}$ & $(2.97{\scriptstyle\pm1.65}){\scriptstyle\times10^{-1}}$ \\
        FNO & $(5.31{\scriptstyle\pm1.17}){\scriptstyle\times10^{-3}}$ & $(2.19{\scriptstyle\pm0.28}){\scriptstyle\times10^{-2}}$ & $(1.15{\scriptstyle\pm0.86}){\scriptstyle\times10^{-1}}$\\
        \bottomrule
    \end{tabular}
    \label{tab:pde}
\end{table}
%%%%%%%%%%%%%%%%%%%%%%%%%%%%%%%%%%%%%%%%%%%%%%%%%%%%%%
\begin{figure}[t]
    \centering
    \includegraphics[width=0.85\columnwidth]{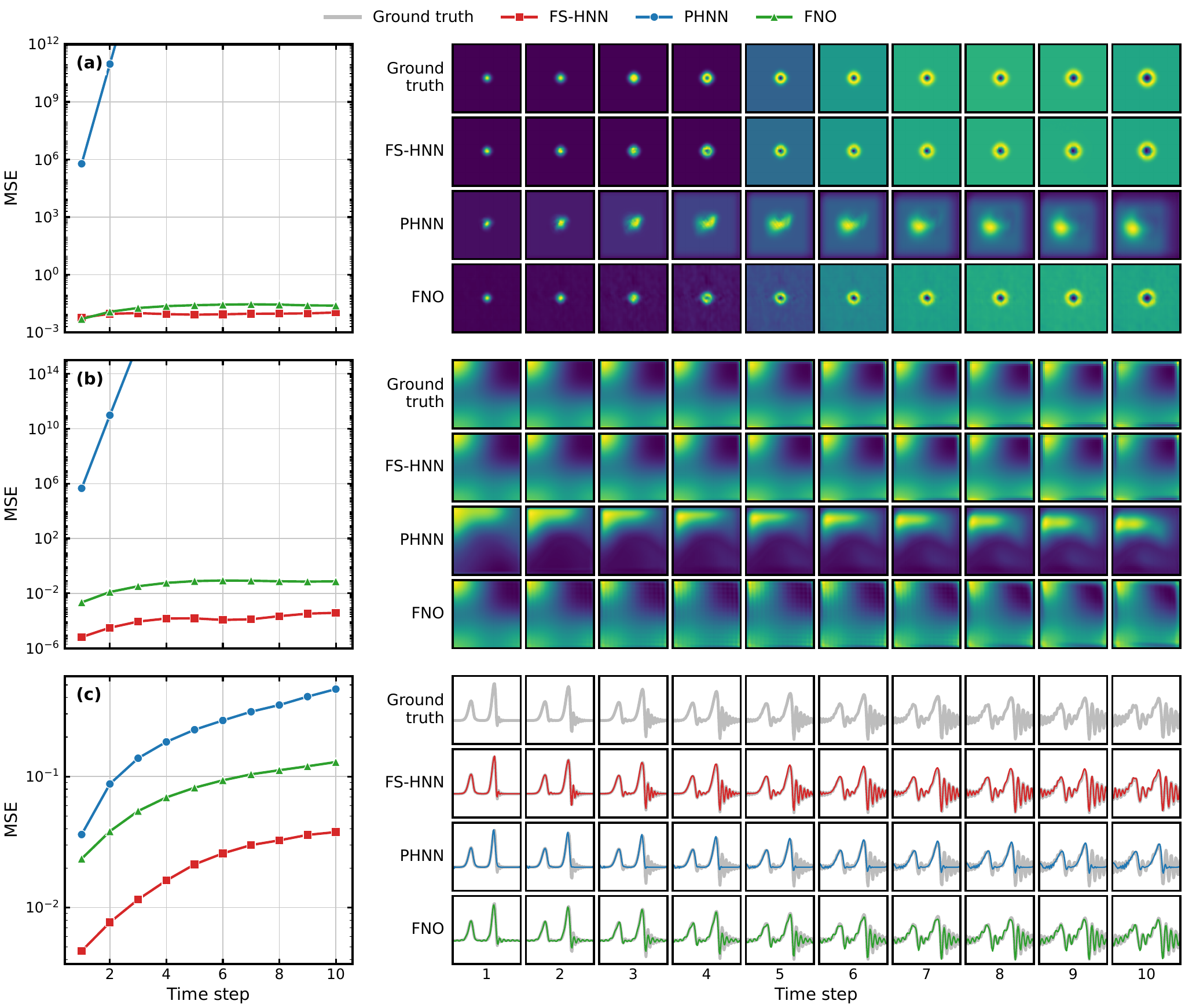}
    \vskip -0.04in
    \caption{\textbf{PDE system rollout comparison.} We compare FS-HNN, PHNN, and FNO on \textbf{(a)} SWE with Gaussian-pulse initialization, \textbf{(b)} SWE with random initialization, and \textbf{(c)} the KdV equation. FS-HNN most closely reproduces the reference fields and achieves the lowest rollout error.}
    \label{fig:pde}
    \vspace{-0.1in}
\end{figure}

\begin{table}[t]
    \centering
    \caption{\textbf{Ablation on nonlinear aggregation.}
    We compare a linear and MLP aggregation. Nonlinear aggregation improves performance on the main nonlinear ODE benchmarks.}
    \label{tab:comb_ablation}
    \vskip 0.03in
    \begin{tabular}{l*{2}{r@{}l}}
        \toprule
        System & \multicolumn{2}{c}{Linear} & \multicolumn{2}{c}{MLP} \\
        \midrule
        Pendulum & $1.01$&${}\times10^{-6}$ & $\mathbf{3.25}$&$\mathbf{{}\times10^{-7}}$ \\
        Double Pendulum & $6.99$&${}\times10^{-2}$ & $\mathbf{3.73}$&$\mathbf{{}\times10^{-2}}$ \\
        FPUT & $3.47$&${}\times10^{-3}$ & $\mathbf{2.10}$&$\mathbf{{}\times10^{-5}}$ \\
        Fast--Slow & $1.26$&${}\times10^{0}$ & $\mathbf{1.45}$&$\mathbf{{}\times10^{-2}}$ \\
        \bottomrule
    \end{tabular}
    \vspace{-0.05in}
\end{table}

\begin{table}[b]
    \vspace{-0.05in}
    \centering
    \caption{\textbf{Ablation on distinct temporal scales.}
    All control models use the same multi-branch architecture and aggregation module, but all branches operate at the same temporal interval. The last row is the full multiscale FS-HNN using multiple temporal intervals.}
    \vskip 0.03in
    \begin{tabular}{lcccc}
        \toprule
        Setting & Pendulum & Double Pendulum & FPUT & Fast--slow \\
        \midrule
        Low frequency only & $2.61\times10^{-6}$ & $3.83\times10^{-2}$ & $2.78\times10^{-3}$ & $1.25\times10^{0}$ \\
        Mid frequency only & $3.53\times10^{-3}$ & $7.23\times10^{-2}$ & $3.42\times10^{-3}$ & $1.80\times10^{2}$ \\
        High frequency only & $8.95\times10^{-1}$ & $3.59\times10^{-1}$  & $3.47\times10^{-3}$ & $2.00\times10^{2}$ \\
        FS-HNN        & $\mathbf{3.25\times10^{-7}}$ & $\mathbf{3.73\times10^{-2}}$ & $\mathbf{2.10\times10^{-5}}$ & $\mathbf{1.45\times10^{-2}}$ \\
        \bottomrule
    \end{tabular}
    \label{tab:same_scale}
\end{table}
%%%%%%%%%%%%%%%%%%%%%%%%%%%%%%%%%%%%%%%%%%%%%%%%%%%%%%
\subsection{PDE Systems}
Section~\ref{subsec:method_pde} extends FS-HNN to PDEs through a learned spatial operator $\mathbf{J}_{\theta}$ and orthogonal projection. We evaluate this construction on the 2D shallow water equations (SWE)~\citep{vreugdenhil2013numerical} with Gaussian-pulse and random initializations, and on the one-dimensional Korteweg--de Vries (KdV) equation. The SWE configurations are inviscid and unforced, so they conserve energy in the continuous setting~\citep{tadmor2008energy}. The random initializations produce more complex spatial structure than the localized Gaussian pulse and are typically harder to learn~\citep{brecht2025physics}. KdV follows the two-soliton setup of Eidnes and Lye~\cite{eidnes2024pseudo}, with $u_{t}+6uu_{x}-u_{xxx}=0$ solved on the periodic domain $[0,20]$. Data generation for both systems is detailed in Appendices~\ref{app:swe} and~\ref{app:kdv}. Models are trained on the first 5 steps and rolled out for the subsequent 50 steps. We compare with PHNN~\citep{eidnes2024pseudo} as the closest structure-preserving baseline, together with Fourier Neural Operator (FNO)~\citep{li2020fourier} and Energy-Consistent Neural Operator (ENO)~\citep{tanaka2025energy}. Rollout accuracy is reported in Table~\ref{tab:pde} and Figure~\ref{fig:pde}..

\textbf{Rollout accuracy.}
Across the systems studied, FS-HNN attains the lowest rollout MSE on all three benchmarks (Table~\ref{tab:pde}). Under both SWE initializations, FS-HNN closely reproduces the reference fields (Figure~\ref{fig:pde}a,b). Its advantage over PHNN is larger for the random initializations, suggesting that the learned spatial operator remains effective as the field dynamics become more complex. On KdV, FS-HNN reproduces the soliton interaction and the dispersive tail (Figure~\ref{fig:pde}c), so the PDE formulation also captures nonlinear wave interactions and dispersive evolution beyond the shallow water setting. These results show that FS-HNN extends frequency-structured Hamiltonian learning to PDE systems without requiring their dynamics operator in closed form.

\textbf{Energy behavior across systems.}
The projection in Section~\ref{subsec:method_pde} is designed to preserve the learned Hamiltonian, which need not coincide with the physical energy. We therefore evaluate the analytic SWE energy and KdV Hamiltonian on the predicted fields and compare them with the reference trajectories. FS-HNN follows the reference energy more closely than PHNN and FNO across the PDE benchmarks, including the small numerical energy variation in the SWE data (Figure~\ref{fig:pdeenergy} and Table~\ref{tab:pde_energy_stats} in Appendix~\ref{pderesults}). As in the ODE setting, FS-HNN rollouts remain faithful to the energy of each system as well as to its state.
%%%%%%%%%%%%%%%%%%%%%%%%%%%%%%%%%%%%%%%%%%%%%%%%%%%%%%
\subsection{Ablation Studies}\label{subsec:ablation}
The ablations below isolate the components of FS-HNN behind these gains. In the ODE setting they examine the aggregator, the use of distinct temporal scales, the Hamiltonian structure, and the cutoffs, and in the PDE setting they separate the contributions of frequency structure and energy preservation.

\textbf{Nonlinear aggregation.} Table~\ref{tab:comb_ablation} compares the MLP aggregator $\mathcal{M}_{\phi}$ with linear aggregation of the same per-band Hamiltonians. Nonlinear aggregation lowers rollout error on all ODE benchmarks, with the largest gains on FPUT and the fast--slow oscillator, where fast and slow dynamics are explicitly coupled. This is consistent with the motivation for $\mathcal{M}_{\phi}$ in Section~\ref{subsec:method_freq_ham}, since coupling breaks the additivity of the per-band Hamiltonians.

\textbf{Distinct temporal scales.} Table~\ref{tab:same_scale} compares FS-HNN with controls that keep its multi-branch architecture and aggregator but filter every branch at the same cutoff, so that capacity is matched and only the distinct temporal scales are removed. These controls underperforms FS-HNN on all four systems, so the improvement comes from combining views at distinct cutoffs rather than from added capacity or the aggregation network.
%%%%%%%%%%%%%%%%%%%%%%%%%%%%%%%%%%%%%%%%%%%%%%%%%%%%%%
\begin{table}[t]
    \centering
    \caption{\textbf{NeuralODE trained with frequency-filtered inputs.} We compare a single-scale NeuralODE with a combined NeuralODE using the same multi-timescale branch aggregation strategy. The combined model gives similar errors on the pendulum and FPUT systems and higher errors on the double pendulum and fast--slow oscillator.}
    \vskip 0.03in
    \begin{tabular}{lcc}
        \toprule
        System & Single & Combined\\
        \midrule
        Pendulum & $1.20\times10^{-9}$ & $\mathbf{1.16\times10^{-9}}$ \\
        Double Pendulum & $\mathbf{1.66\times10^{-2}}$ & $1.52\times10^{-1}$ \\
        FPUT & $3.47\times10^{-3}$ & $\mathbf{3.44\times10^{-3}}$ \\
        Fast--slow oscillator & $\mathbf{4.07\times10^{-2}}$ & $7.83\times10^{-2}$ \\
        \bottomrule
    \end{tabular}
    \label{tab:neuralode_multiscale}
    \vspace{-0.05in}
\end{table}
%%%%%%%%%%%%%%%%%%%%%%%%%%%%%%%%%%%%%%%%%%%%%%%%%%%%%%

\textbf{Hamiltonian structure.} Applying the same frequency-filtered views and aggregation to an unconstrained NeuralODE~\citep{chen2018neural} isolates the role of Hamiltonian structure. The combined NeuralODE gives errors similar to its single-scale version on the pendulum and FPUT and higher errors on the double pendulum and fast--slow oscillator (Table~\ref{tab:neuralode_multiscale}), while FS-HNN benefits from combining its components. Each FS-HNN branch contributes a scalar Hamiltonian and aggregation yields a single Hamiltonian with canonical dynamics, whereas NeuralODE combines unconstrained vector fields. Together with the previous ablation, this comparison supports combining information across temporal scales at the level of the Hamiltonian rather than the vector field.
%%%%%%%%%%%%%%%%%%%%%%%%%%%%%%%%%%%%%%%%%%%%%%%%%%%%%%
%%%%%%%%%%%%%%%%%%%%%%%%%%%%%%%%%%%%%%%%%%%%%%%%%%%%%%

\textbf{Number of views and cutoff placement.} The number of nested views $K$ and their cutoffs determine which temporal scales the views expose. Tables~\ref{tab:ksweep} and~\ref{tab:cutoffs} in Appendix~\ref{oderesults} vary $K$ and compare fixed cutoffs with cutoffs selected from the observed spectrum. Adding nested views improves accuracy on FPUT, whereas on the fast--slow oscillator, aligning the cutoffs with the system’s spectral scales is more effective than adding views with cutoffs obtained by repeated halving. Spectrum-selected cutoffs with $K=3$ give the lowest error of all tested configurations on both systems. The same choice also outperforms the fixed cutoffs on the remaining ODE benchmarks and on KdV.

\textbf{Frequency structure and energy preservation.} The PDE construction combines frequency structure, through the filtered views and aggregation, with energy preservation, through the learned operator and projection. Either ingredient could account for the gains in Table~\ref{tab:pde}. Table~\ref{tab:pde_components} in Appendix~\ref{pderesults} separates them on random-initialization SWE and KdV with two controls, each removing one ingredient. The single-scale conservative control removes the frequency structure and keeps the DeepONet Hamiltonian, learned $\mathbf{J}_{\theta}$, and projection. The non-conservative DeepONet removes energy preservation and keeps the frequency structure. FS-HNN gives lower rollout error than either control on both benchmarks, so its PDE gains come from combining frequency structure with energy preservation rather than from either ingredient alone.
%%%%%%%%%%%%%%%%%%%%%%%%%%%%%%%%%%%%%%%%%%%%%%%%%%%%%%
%%%%%%%%%%%%%%%%%%%%%%%%%%%%%%%%%%%%%%%%%%%%%%%%%%%%%%

%%%%%%%%%%%%%%%%%%%%%%%%%%%%%%%%%%%%%%%%%%%%%%%
\section{Conclusion and Discussion}
%%%%%%%%%%%%%%%%%%%%%%%%%%%%%%%%%%%%%%%%%%%%%%%%%%%%%%%%%%
Hamiltonian neural networks build energy conservation into the learned dynamics but inherit spectral bias, which can leave fast oscillations poorly resolved. FS-HNN addresses this by organizing the Hamiltonian around the frequency structure of the observed trajectories. Nested frequency-filtered views supervise per-band Hamiltonian components, and a learned nonlinear aggregator combines them into a single scalar Hamiltonian whose dynamics capture interactions between fast and slow modes and conserve the learned energy by construction. For PDEs, FS-HNN represents the components as neural functionals, learns the action of the dynamics operator, and conserves the learned energy through orthogonal projection. Across the ODE and PDE benchmarks, FS-HNN attains lower rollout error than black-box and structure-preserving baselines while remaining faithful to the physical energy of each system. Ablations attribute these gains to combining distinct temporal scales at the level of a scalar Hamiltonian rather than to added capacity, and show that frequency structure and energy preservation are complementary for PDEs. These results suggest that structure-preserving models benefit from dedicating part of the model to coarser temporal structure rather than learning from the full-spectrum trajectory alone.

\textbf{Scope and limitations.} The PDE experiments focus on smooth conservative dynamics on fixed spatial grids. Shock-dominated or strongly dissipative regimes require entropy-aware or non-conservative formulations. Conservation of the learned Hamiltonian holds in continuous time, so exact conservation after time discretization depends on the integrator, and the projected PDE dynamics are not guaranteed to be symplectic. Filtering a Hamiltonian trajectory does not in general produce a Hamiltonian flow, so each branch fits a Hamiltonian approximation, and the structure guarantees apply to the aggregated dynamics rather than to the individual branches. The projection also adds computational cost through gradient evaluations. Extending the implementation across spatial resolutions and boundary conditions remains future work.
%%%%%%%%%%%%%%%%%%%%%%%%%%%%%%%%%%%%%%%%%%%%%%%%%%%%%%%%%%
\begin{ack}
    Yulong Yang was supported by a grant from the National Aeronautics and Space Administration funded through the Jet Propulsion Laboratory Strategic University Research Partnerships program. We sincerely acknowledge the NeurIPS 2026 reviewers and area chair for their thoughtful feedback. The authors declare no competing interests.
\end{ack}
%%%%%%%%%%%%%%%%%%%%%%%%%%%%%%%%%%%%%%%%%%%%%%%%%%%%%%%%%%
%%%%%%%%%%%%%%%%%%%%%%%%%%%%%%%%%%%%%%%%%%%%%%%
\bibliography{ref}
\bibliographystyle{unsrt}
%%%%%%%%%%%%%%%%%%%%%%%%%%%%%%%%%%%%%%%%%%%%%%%
\newpage
\appendix
\section{Hamiltonian Structure and Conservation}\label{app:theory}
The derivations below expand on the Hamiltonian and multi-timescale structure used in the main text. Hamilton’s equations and the HNN objective are restated in the notation used throughout. The slow--fast splitting of a stiff Hamiltonian makes precise the separation of characteristic frequencies that motivates the frequency-filtered views. The conservation analysis establishes that the orthogonal projection conserves learned Hamiltonians in continuous time and admits a skew-gradient form, and explains why each branch fits a Hamiltonian approximation rather than an exact Hamiltonian flow.

\subsection{Hamiltonian Neural Networks}
For a conservative Hamiltonian system, we write the state at time \(t\) as
\[
\mathbf{z}(t)=\begin{bmatrix} \mathbf{q}(t) \\ \mathbf{p}(t) \end{bmatrix}\in\mathbb{R}^{2d},
\]
where \(\mathbf{q}(t)\in\mathbb{R}^d\) and \(\mathbf{p}(t)\in\mathbb{R}^d\) are the generalized coordinates and momenta. The scalar Hamiltonian \(\mathcal{H}(\mathbf{q},\mathbf{p})\) typically represents the total energy.

The dynamics follow Hamilton's equations,
\begin{equation}
\dot{\mathbf{q}}=\frac{\partial \mathcal{H}}{\partial \mathbf{p}}, 
\qquad 
\dot{\mathbf{p}}=-\frac{\partial \mathcal{H}}{\partial \mathbf{q}}.
\label{eq:hamilton_equations}
\end{equation}

To write these equations compactly, we introduce the canonical symplectic matrix
\begin{equation}
\mathbf{J}=
\begin{bmatrix}
0 & \mathbf{I} \\
-\mathbf{I} & 0
\end{bmatrix},
\qquad \mathbf{J}^{\intercal}=-\mathbf{J},
\label{eq:canonical_J}
\end{equation}
where \(\mathbf{I}\) is the \(d\times d\) identity matrix. Writing the state gradient as

\[
\boldsymbol{\nabla}_{\mathbf{z}} \mathcal{H}(\mathbf{z})=
\begin{bmatrix}
\frac{\partial \mathcal{H}}{\partial \mathbf{q}}(\mathbf{q},\mathbf{p})\\[2pt]
\frac{\partial \mathcal{H}}{\partial \mathbf{p}}(\mathbf{q},\mathbf{p})
\end{bmatrix},
\]

Hamilton's equations take the canonical form
\begin{equation}
\dot{\mathbf{z}}=\mathbf{J}\boldsymbol{\nabla}_{\mathbf{z}} \mathcal{H}(\mathbf{z}).
\label{eq:symplectic_form}
\end{equation}
We can check energy conservation by differentiating the Hamiltonian along a trajectory,
\begin{equation}
\frac{d\mathcal{H}}{dt}
=\boldsymbol{\nabla}_{\mathbf{z}} \mathcal{H}(\mathbf{z})^{\intercal} \dot{\mathbf{z}}.
\label{eq:energy_chain_rule}
\end{equation}
Substituting \eqref{eq:symplectic_form} into \eqref{eq:energy_chain_rule} gives
\begin{equation}
\frac{d\mathcal{H}}{dt}
=\boldsymbol{\nabla}_{\mathbf{z}} \mathcal{H}(\mathbf{z})^{\intercal} \mathbf{J} \boldsymbol{\nabla}_{\mathbf{z}} \mathcal{H}(\mathbf{z}).
\label{eq:energy_substitution}
\end{equation}
Since \(\mathbf{J}^{\intercal}=-\mathbf{J}\), the quadratic form \(\mathbf{x}^{\intercal} \mathbf{J} \mathbf{x}\) vanishes for any vector \(\mathbf{x}\), and therefore
\begin{equation}
\frac{d\mathcal{H}}{dt}=0.
\label{eq:energy_conservation}
\end{equation}
The continuous-time trajectories therefore remain on level sets of the Hamiltonian.

A Hamiltonian Neural Network (HNN) represents the unknown Hamiltonian with a neural network \(\mathcal{H}_{\theta}(\mathbf{z})\) and defines the vector field by
\begin{equation}
\dot{\mathbf{z}}_\theta = \mathbf{J} \boldsymbol{\nabla}_{\mathbf{z}} \mathcal{H}_{\theta}(\mathbf{z}).
\label{eq:hnn_dynamics}
\end{equation}
Given training data \(\{(\mathbf{z}_i,\dot{\mathbf{z}}_i)\}_{i=1}^N\), the parameters \(\theta\) are typically learned by minimizing
\begin{equation}
\mathcal{L}(\theta)
=\frac{1}{N}\sum_{i=1}^N
\left\|
\dot{\mathbf{z}}_i - \mathbf{J}\boldsymbol{\nabla}_{\mathbf{z}} \mathcal{H}_{\theta}(\mathbf{z}_i)
\right\|_2^2.
\label{eq:hnn_loss}
\end{equation}
This parameterization enforces Hamiltonian structure and conserves the learned Hamiltonian in continuous time.

\subsection{Slow--Fast Splitting in Hamiltonian Form}\label{app:slowfast}
Following the multirate setting in \cite{ober2024variational}, assume the potential energy can be additively decomposed into slow and fast parts,
\begin{equation}
U(\mathbf{q})=V(\mathbf{q})+W(\mathbf{q}),\qquad W(\mathbf{q})=\frac{1}{\varepsilon}\,\widetilde{W}(\mathbf{q}),\quad 0<\varepsilon\ll 1.
\label{eq:U_split_H}
\end{equation}

The scaled potential $W$ generates stiff forces. Split the configuration into slow and fast variables
\begin{equation}
\mathbf{q}=\begin{bmatrix} \mathbf{q}^{s} \\ \mathbf{q}^{f} \end{bmatrix},
\qquad 
\mathbf{q}^{s}\in Q_s,\; \mathbf{q}^{f}\in Q_f,\; Q_s\times Q_f = Q,
\label{eq:q_split_H}
\end{equation}
and assume (i) the fast potential depends only on the fast variables, $W=W(\mathbf{q}^{f})$, while the slow potential may depend on all variables, $V=V(\mathbf{q}^{s},\mathbf{q}^{f})$, and (ii) the kinetic energy is quadratic with a block-diagonal mass matrix,
\begin{equation}
T(\dot{\mathbf{q}})=\frac{1}{2}\dot{\mathbf{q}}^{\intercal} \mathbf{M}\dot{\mathbf{q}},
\qquad 
\mathbf{M}=\begin{bmatrix} \mathbf{M}_s & 0 \\ 0 & \mathbf{M}_f \end{bmatrix}.
\label{eq:kinetic_block}
\end{equation}
Under these assumptions, the Hamiltonian admits the slow--fast decomposition used in the multirate formulation below.

We first write the corresponding Lagrangian as
\begin{equation}
L(\mathbf{q}^{s},\mathbf{q}^{f},\dot{\mathbf{q}}^{s},\dot{\mathbf{q}}^{f})
=\frac12 (\dot{\mathbf{q}}^{s})^{\intercal} \mathbf{M}_s \dot{\mathbf{q}}^{s}+\frac12 (\dot{\mathbf{q}}^{f})^{\intercal} \mathbf{M}_f \dot{\mathbf{q}}^{f} - V(\mathbf{q}^{s},\mathbf{q}^{f})-W(\mathbf{q}^{f}).
\label{eq:L_split}
\end{equation}

Define the conjugate momenta by the Legendre map
\begin{equation}
\mathbf{p}^{s} := \frac{\partial L}{\partial \dot{\mathbf{q}}^{s}}=\mathbf{M}_s\dot{\mathbf{q}}^{s},
\qquad
\mathbf{p}^{f} := \frac{\partial L}{\partial \dot{\mathbf{q}}^{f}}=\mathbf{M}_f\dot{\mathbf{q}}^{f},
\label{eq:momenta_def}
\end{equation}

The velocities are therefore
\begin{equation}
\dot{\mathbf{q}}^{s} = \mathbf{M}_s^{-1}\mathbf{p}^{s},
\qquad
\dot{\mathbf{q}}^{f} = \mathbf{M}_f^{-1}\mathbf{p}^{f}.
\label{eq:velocities_from_p}
\end{equation}

The Legendre transform gives
\begin{equation}
\mathcal{H}(\mathbf{q}^{s},\mathbf{q}^{f},\mathbf{p}^{s},\mathbf{p}^{f})=(\mathbf{p}^{s})^{\intercal} \dot{\mathbf{q}}^{s} + (\mathbf{p}^{f})^{\intercal} \dot{\mathbf{q}}^{f} - L(\mathbf{q}^{s},\mathbf{q}^{f},\dot{\mathbf{q}}^{s},\dot{\mathbf{q}}^{f}).
\label{eq:legendre}
\end{equation}

Here, the velocities $\dot{\mathbf{q}}^{s},\dot{\mathbf{q}}^{f}$ are expressed in terms of $\mathbf{p}^{s},\mathbf{p}^{f}$ using \eqref{eq:velocities_from_p}. Substituting \eqref{eq:L_split}--\eqref{eq:velocities_from_p} gives
\begin{equation}
\mathcal{H}(\mathbf{q}^{s},\mathbf{q}^{f},\mathbf{p}^{s},\mathbf{p}^{f})
=\frac12 (\mathbf{p}^{s})^{\intercal} \mathbf{M}_s^{-1}\mathbf{p}^{s}+\frac12 (\mathbf{p}^{f})^{\intercal} \mathbf{M}_f^{-1}\mathbf{p}^{f}+V(\mathbf{q}^{s},\mathbf{q}^{f})+W(\mathbf{q}^{f}).
\label{eq:H_split}
\end{equation}

Hamilton's equations are
\begin{equation}
\dot{\mathbf{q}}^{s}=\frac{\partial \mathcal{H}}{\partial \mathbf{p}^{s}},\quad
\dot{\mathbf{p}}^{s}=-\frac{\partial \mathcal{H}}{\partial \mathbf{q}^{s}},\quad
\dot{\mathbf{q}}^{f}=\frac{\partial \mathcal{H}}{\partial \mathbf{p}^{f}},\quad
\dot{\mathbf{p}}^{f}=-\frac{\partial \mathcal{H}}{\partial \mathbf{q}^{f}}.
\label{eq:Ham_general}
\end{equation}

Applying \eqref{eq:H_split} yields the split system
\begin{equation}
\dot{\mathbf{q}}^{s} = \mathbf{M}_s^{-1}\mathbf{p}^{s},
\qquad
\dot{\mathbf{p}}^{s} = -\frac{\partial V}{\partial \mathbf{q}^{s}}(\mathbf{q}^{s},\mathbf{q}^{f}),
\label{eq:Ham_slow}
\end{equation}
\begin{equation}
\dot{\mathbf{q}}^{f} = \mathbf{M}_f^{-1}\mathbf{p}^{f},
\qquad
\dot{\mathbf{p}}^{f} = -\frac{\partial V}{\partial \mathbf{q}^{f}}(\mathbf{q}^{s},\mathbf{q}^{f}) - \boldsymbol{\nabla}_{\mathbf{q}^{f}} W(\mathbf{q}^{f}),
\label{eq:Ham_fast}
\end{equation}

Next, substituting the stiffness scaling \eqref{eq:U_split_H}, with $W(\mathbf{q}^{f})=\varepsilon^{-1}\bar W(\mathbf{q}^{f})$, gives
\begin{equation}
\boldsymbol{\nabla}_{\mathbf{q}^{f}} W(\mathbf{q}^{f})=\frac{1}{\varepsilon}\boldsymbol{\nabla}_{\mathbf{q}^{f}} \bar W(\mathbf{q}^{f}),
\end{equation}
so the fast momentum equation becomes
\begin{equation}
\dot{\mathbf{p}}^{f}
=
-\frac{\partial V}{\partial \mathbf{q}^{f}}(\mathbf{q}^{s},\mathbf{q}^{f})
-\frac{1}{\varepsilon}\boldsymbol{\nabla}_{\mathbf{q}^{f}} \bar W(\mathbf{q}^{f}),
\label{eq:fast_stiff_force}
\end{equation}
so the fast subsystem contains an $\mathcal{O}(\varepsilon^{-1})$ force, whereas the slow subsystem in \eqref{eq:Ham_slow} contains no such stiff term.

To identify the fast timescale, let $\mathbf{q}_*^{f}$ be a local minimizer of $\bar W$ with $\boldsymbol{\nabla}_{\mathbf{q}^{f}} \bar W(\mathbf{q}_*^{f})=0$, and define $\boldsymbol{\eta} := \mathbf{q}^{f}-\mathbf{q}_*^{f}$.
Linearizing $\boldsymbol{\nabla}_{\mathbf{q}^{f}} \bar W(\mathbf{q}^{f})$ near $\mathbf{q}_*^{f}$ gives $\boldsymbol{\nabla}_{\mathbf{q}^{f}} \bar W(\mathbf{q}^{f})\approx \mathbf{K}_f\boldsymbol{\eta}$ with $\mathbf{K}_f=\boldsymbol{\nabla}_{\mathbf{q}^{f}}^2 \bar W(\mathbf{q}_*^{f})\succeq 0$.
Ignoring lower-order coupling terms in $\partial V/\partial \mathbf{q}^{f}$ relative to $\varepsilon^{-1}$, we combine
\[
\dot{\mathbf{q}}^{f} = \mathbf{M}_f^{-1}\mathbf{p}^{f},
\qquad
\dot{\mathbf{p}}^{f} \approx -\frac{1}{\varepsilon}\mathbf{K}_f\boldsymbol{\eta}
\]
to obtain the leading fast oscillator
\begin{equation}
\ddot{\boldsymbol{\eta}} \approx -\frac{1}{\varepsilon}\mathbf{M}_f^{-1}\mathbf{K}_f\,\boldsymbol{\eta}.
\label{eq:fast_oscillator_H_app}
\end{equation}
The fast frequencies thus scale as $\omega \sim \varepsilon^{-1/2}$, giving a shorter characteristic timescale for $\mathbf{q}^{f}$ than for $\mathbf{q}^{s}$.

Finally, collecting the variables into the phase-space vector
\[
\mathbf{z} := \begin{bmatrix} \mathbf{q}^{s} & \mathbf{q}^{f} & \mathbf{p}^{s} & \mathbf{p}^{f} \end{bmatrix}^{\intercal},
\]
we can write the system in \eqref{eq:Ham_general} in canonical symplectic form
\begin{equation}
\dot{\mathbf{z}} = \mathbf{J} \boldsymbol{\nabla}_{\mathbf{z}} \mathcal{H}(\mathbf{z}),
\qquad
\mathbf{J}=
\begin{bmatrix}
0 & 0 & \mathbf{I} & 0\\
0 & 0 & 0 & \mathbf{I}\\
-\mathbf{I}& 0 & 0 & 0\\
0 & -\mathbf{I}& 0 & 0
\end{bmatrix},
\label{eq:canonical_symplectic_split}
\end{equation}
where the block sizes match $(\mathbf{q}^{s},\mathbf{q}^{f},\mathbf{p}^{s},\mathbf{p}^{f})$.

If the slow potential depends only on the slow variables, $V=V(\mathbf{q}^{s})$, the block-diagonal kinetic energy in \eqref{eq:kinetic_block} allows the Hamiltonian to separate as
\[
\mathcal{H}(\mathbf{z})=\mathcal{H}_s(\mathbf{q}^{s},\mathbf{p}^{s})+\mathcal{H}_f(\mathbf{q}^{f},\mathbf{p}^{f}),
\]
and the slow and fast Hamiltonian subsystems evolve independently.

\subsection{Orthogonality constraint versus strict conservation.}\label{app:symplectic}

FS-HNN imposes orthogonality on the learned vector field. It does not directly require
\begin{equation}
    \mathcal{H}(\mathbf{z}_{t+1}) = \mathcal{H}(\mathbf{z}_t)
\end{equation}
at each rollout step or enforce exact symplectic or multisymplectic preservation.
The orthogonality condition makes the continuous-time vector field tangent to
the level sets of the Hamiltonian.

Let $\mathcal{H}(\mathbf{z})$ denote the learned Hamiltonian and let $\mathbf{f}_\theta(\mathbf{z})$ denote the
unconstrained vector-field prediction before projection. We impose the condition
\begin{equation}
    \boldsymbol{\nabla}_{\mathbf{z}} \mathcal{H}(\mathbf{z})^{\intercal} \mathbf{f}(\mathbf{z}) = 0,
    \label{eq:orthogonal_condition}
\end{equation}
where $\mathbf{f}(\mathbf{z})$ is the projected vector field. This condition implies that, in
continuous time,
\begin{equation}
    \frac{d\mathcal{H}}{dt}
    =
    \boldsymbol{\nabla}_{\mathbf{z}} \mathcal{H}(\mathbf{z})^{\intercal} \dot{\mathbf{z}}
    =
    \boldsymbol{\nabla}_{\mathbf{z}} \mathcal{H}(\mathbf{z})^{\intercal} \mathbf{f}(\mathbf{z})
    =
    0.
    \label{eq:continuous_energy_preservation}
\end{equation}
This continuous-time conservation property does not guarantee exact conservation
after numerical time integration.

The orthogonality condition also gives a skew-gradient representation. At points where
$\boldsymbol{\nabla}_{\mathbf{z}} \mathcal{H}(\mathbf{z})\neq 0$, if
\begin{equation}
    \boldsymbol{\nabla}_{\mathbf{z}} \mathcal{H}(\mathbf{z})^{\intercal} \mathbf{f}(\mathbf{z})=0,
\end{equation}
then there exists a skew-symmetric structure operator $\mathbf{J}(\mathbf{z})$ such that
\begin{equation}
    \mathbf{f}(\mathbf{z})=\mathbf{J}(\mathbf{z})\boldsymbol{\nabla}_{\mathbf{z}} \mathcal{H}(\mathbf{z}).
\end{equation}
One way to see this is to construct
\begin{equation}
    \mathbf{J}(\mathbf{z})
    =
    \frac{
        \mathbf{f}(\mathbf{z})\boldsymbol{\nabla}_{\mathbf{z}} \mathcal{H}(\mathbf{z})^{\intercal}
        -
        \boldsymbol{\nabla}_{\mathbf{z}} \mathcal{H}(\mathbf{z})\mathbf{f}(\mathbf{z})^{\intercal}
    }{
        \|\boldsymbol{\nabla}_{\mathbf{z}} \mathcal{H}(\mathbf{z})\|_2^2
    }.
    \label{eq:constructed_skew_operator}
\end{equation}
This operator satisfies
\begin{equation}
    \mathbf{J}(\mathbf{z})^{\intercal} = -\mathbf{J}(\mathbf{z}),
\end{equation}
and, using the orthogonality condition
$\boldsymbol{\nabla}_{\mathbf{z}} \mathcal{H}(\mathbf{z})^{\intercal} \mathbf{f}(\mathbf{z})=0$, we obtain
\begin{equation}
\begin{aligned}
    \mathbf{J}(\mathbf{z})\boldsymbol{\nabla}_{\mathbf{z}} \mathcal{H}(\mathbf{z})
    &=
    \frac{
        \mathbf{f}(\mathbf{z})\boldsymbol{\nabla}_{\mathbf{z}} \mathcal{H}(\mathbf{z})^{\intercal} \boldsymbol{\nabla}_{\mathbf{z}} \mathcal{H}(\mathbf{z})
        -
        \boldsymbol{\nabla}_{\mathbf{z}} \mathcal{H}(\mathbf{z})\mathbf{f}(\mathbf{z})^{\intercal} \boldsymbol{\nabla}_{\mathbf{z}} \mathcal{H}(\mathbf{z})
    }{
        \|\boldsymbol{\nabla}_{\mathbf{z}} \mathcal{H}(\mathbf{z})\|_2^2
    } \\
    &=
    \frac{
        \mathbf{f}(\mathbf{z})\|\boldsymbol{\nabla}_{\mathbf{z}} \mathcal{H}(\mathbf{z})\|_2^2
        -
        \boldsymbol{\nabla}_{\mathbf{z}} \mathcal{H}(\mathbf{z})\big(\mathbf{f}(\mathbf{z})^{\intercal} \boldsymbol{\nabla}_{\mathbf{z}} \mathcal{H}(\mathbf{z})\big)
    }{
        \|\boldsymbol{\nabla}_{\mathbf{z}} \mathcal{H}(\mathbf{z})\|_2^2
    } \\
    &=
    \mathbf{f}(\mathbf{z}).
\end{aligned}
\end{equation}
Under exact orthogonality, the learned vector field therefore admits a
skew-symmetric structure operator away from critical points of the Hamiltonian.
This establishes energy preservation in the continuous-time skew-gradient sense,
but does not establish symplectic or multisymplectic structure.

The learned functional need not coincide with the physical energy. Away from critical points, a system with a first integral $I$ admits a local representation $\dot{\mathbf{z}}=\mathbf{J}(\mathbf{z})\boldsymbol{\nabla}_{\mathbf{z}}I(\mathbf{z})$ with a non-unique skew-symmetric $\mathbf{J}(\mathbf{z})$~\cite{mclachlan1999geometric}. FS-HNN learns such an invariant--operator pair from the observed flow. Identifying the learned $\mathcal{H}$ with the analytic energy, or recovering the physical Poisson operator, requires additional physical priors or direct supervision~\cite{bridges1997multi,bridges2001multi}.

\textbf{Hamiltonian approximation of filtered trajectories.}
The frequency-filtered trajectories provide training targets at different temporal resolutions. Filtering does not generally produce exact Hamiltonian flows, so each branch fits a Hamiltonian approximation to its target. The aggregated scalar Hamiltonian then defines the final ODE vector field through the canonical symplectic matrix.

\section{Data Generation}
The subsections below specify the governing equations, numerical integrator, and initial conditions of each benchmark, covering the four ODE systems, the two shallow water configurations, and KdV. Because every reference trajectory comes from numerical integration, its energy is conserved only up to discretization error, so the physical-energy error reported in the main text is measured against the energy of the reference trajectory rather than a constant.

\subsection{Pendulum.}\label{app:pendulum}
For the simple pendulum, we write the state as
\(\mathbf{z}=(\boldsymbol{\theta},\boldsymbol{\omega})^{{\intercal}}\), with dynamics
\[
\dot{\mathbf{z}}
=
\begin{bmatrix}
\dot{\boldsymbol{\theta}}\\
\dot{\boldsymbol{\omega}}
\end{bmatrix}
=
\begin{bmatrix}
\boldsymbol{\omega}\\
-(g/L)\sin\boldsymbol{\theta}
\end{bmatrix}.
\]
We generate data using a symplectic (leapfrog) scheme: first update \(\boldsymbol{\omega}\) to a half step,
then advance \(\boldsymbol{\theta}\), and finally update \(\boldsymbol{\omega}\) to the full step
\(\big(\boldsymbol{\omega}_{n+\frac12},\boldsymbol{\theta}_{n+1},\boldsymbol{\omega}_{n+1}\big)\). 
A single random initial condition is sampled with \(\boldsymbol{\theta}_0\sim\mathcal{U}[-\pi,\pi]\) and
\(\boldsymbol{\omega}_0\sim\mathcal{U}[-1,1]\), and the system is integrated for
\(N_{\mathrm{traj}}\times N_{\mathrm{steps}}\) steps, after which the long trajectory is reshaped into
\(N_{\mathrm{traj}}\) segments. 
Optionally, i.i.d.\ Gaussian noise with standard deviation \(\sigma\) can be added to both \(\boldsymbol{\theta}\) and \(\boldsymbol{\omega}\). 
The underlying continuous-time pendulum is conservative (no damping/forcing), and the leapfrog integrator is symplectic, which yields a bounded energy error and avoids systematic energy drift over long horizons.

\subsection{Double Pendulum.}\label{app:doublependulum}
For the double pendulum, we write the state as
\(\mathbf{z}=(\boldsymbol{\theta},\boldsymbol{\omega})^{{\intercal}}\), where
\(\boldsymbol{\theta}=(\boldsymbol{\theta}_1,\boldsymbol{\theta}_2)^{{\intercal}}\) and
\(\boldsymbol{\omega}=(\boldsymbol{\omega}_1,\boldsymbol{\omega}_2)^{{\intercal}}\), with parameters
\((m_1,m_2,L_1,L_2,g)\).
The acceleration vector \(\boldsymbol{\alpha}=(\boldsymbol{\alpha}_1,\boldsymbol{\alpha}_2)^{{\intercal}}\) is computed from the standard coupled equations, so the dynamics can be written as
\[
\dot{\mathbf{z}}
=
\begin{bmatrix}
\dot{\boldsymbol{\theta}}\\
\dot{\boldsymbol{\omega}}
\end{bmatrix}
=
\begin{bmatrix}
\boldsymbol{\omega}\\
\boldsymbol{\alpha}(\boldsymbol{\theta},\boldsymbol{\omega})
\end{bmatrix}.
\]
Time stepping is performed by a semi-implicit \emph{symplectic Euler} update: first update \(\boldsymbol{\omega}\),
then update \(\boldsymbol{\theta}\) using the new velocities.
Angles are wrapped back to \((-\pi,\pi]\) after each update to avoid unbounded growth. 
As in the pendulum case, we sample one random initial condition
\(\boldsymbol{\theta}_{0}=(\boldsymbol{\theta}_{1,0},\boldsymbol{\theta}_{2,0})^{{\intercal}}\) with \(\boldsymbol{\theta}_{1,0},\boldsymbol{\theta}_{2,0}\sim\mathcal{U}[-\pi,\pi]\), and
\(\boldsymbol{\omega}_{0}=(\boldsymbol{\omega}_{1,0},\boldsymbol{\omega}_{2,0})^{{\intercal}}\) with \(\boldsymbol{\omega}_{1,0},\boldsymbol{\omega}_{2,0}\sim\mathcal{U}[-1,1]\).
We evolve the system for \(N_{\mathrm{traj}}\times N_{\mathrm{steps}}\) steps and reshape the long roll-out into
\(N_{\mathrm{traj}}\) trajectory segments. 
Optional additive Gaussian noise is supported for both \(\boldsymbol{\theta}\) and \(\boldsymbol{\omega}\). 
The continuous double pendulum (without dissipation) is energy-conserving, and the symplectic Euler discretization preserves the symplectic structure, typically producing stable long-term behavior with bounded (though not exactly zero) energy error.

\subsection{Fermi-Pasta-Ulam-Tsingou.}\label{app:fermi}
For this 1D particle chain, the state is \((\mathbf{q},\mathbf{p})\in\mathbb{R}^{N}\times\mathbb{R}^{N}\), where \(\mathbf{q}\) and \(\mathbf{p}\) collect the particle displacements and momenta. The vector dynamics are Hamiltonian:
\[
\dot{\mathbf{q}}=\frac{\mathbf{p}}{m},\qquad 
\dot{\mathbf{p}}=\mathbf{F}(\mathbf{q}),
\]
where the \(i\)-th component of the force vector is
\[
\mathbf{F}_i(\mathbf{q})=\mathbf{f}_{i-1}(\mathbf{q})-\mathbf{f}_i(\mathbf{q}).
\]
We use periodic boundary conditions \(\mathbf{q}_{N+1}\equiv \mathbf{q}_1\) and \(\mathbf{q}_0\equiv \mathbf{q}_N\) unless otherwise specified. Let \(\mathbf{r}_i(\mathbf{q})=\mathbf{q}_{i+1}-\mathbf{q}_i\) be the nearest-neighbor stretch. The scalar spring force magnitude associated with bond \(i\) is
\[
\mathbf{f}_i(\mathbf{q})=k\,\mathbf{r}_i+\alpha \mathbf{r}_i^2+\beta \mathbf{r}_i^3.
\]
Equivalently, the system corresponds to the Hamiltonian
\[
\mathcal{H}(\mathbf{q},\mathbf{p})=\sum_{i=1}^{N}\frac{\mathbf{p}_i^2}{2m}+\sum_{i=1}^{N}\left(\frac{k}{2}\mathbf{r}_i^2+\frac{\alpha}{3}\mathbf{r}_i^3+\frac{\beta}{4}\mathbf{r}_i^4\right).
\]
In many experiments we set \(\alpha=0\) and \(\beta>0\).

Data are generated using a symplectic velocity--Verlet integrator. Given \((\mathbf{q}_n,\mathbf{p}_n)\), compute the force \(\mathbf{F}(\mathbf{q}_n)\), update positions
\[
\mathbf{q}_{n+1}=\mathbf{q}_n+\frac{\mathbf{p}_n}{m}\Delta \mathbf{t}+\tfrac12\frac{\mathbf{F}(\mathbf{q}_n)}{m}\Delta \mathbf{t}^2,
\]
recompute \(\mathbf{F}(\mathbf{q}_{n+1})\), and then update momenta
\[
\mathbf{p}_{n+1}=\mathbf{p}_n+\tfrac12\big(\mathbf{F}(\mathbf{q}_n)+\mathbf{F}(\mathbf{q}_{n+1})\big)\Delta \mathbf{t}.
\]
A random initial condition is sampled as
\[
\mathbf{q}_0\sim\mathcal{N}(0,\sigma_q^2 \mathbf{I}),\qquad \mathbf{p}_0\sim\mathcal{N}(0,\sigma_p^2 \mathbf{I}).
\]
The system is integrated for \(N_{\mathrm{traj}}\times N_{\mathrm{steps}}\) steps with fixed \(\Delta \mathbf{t}\), saving every \(s\) steps. The resulting long roll-out is reshaped into \(N_{\mathrm{traj}}\) trajectory segments of length
\[
T=\left\lfloor \frac{N_{\mathrm{steps}}}{s}\right\rfloor+1.
\]
Optionally, i.i.d.\ Gaussian noise with standard deviation \(\sigma\) can be added to both \(\mathbf{q}\) and \(\mathbf{p}\). We also record the total energy \(E=\mathcal{H}(\mathbf{q},\mathbf{p})\) at each saved frame. The underlying continuous-time system is conservative (no damping or forcing), and the velocity--Verlet scheme is symplectic, which typically yields stable long-horizon simulations with bounded energy error rather than systematic energy drift.

\subsection{Fast--Slow Oscillator.}\label{app:fastslow}
We use the nearly periodic Hamiltonian system of Duruisseaux et al.~\cite{duruisseaux2023approximation}, with state \(\mathbf{z}=(q_{1},q_{2},p_{1},p_{2})^{{\intercal}}\) and Hamiltonian
\[
\mathcal{H}_{\varepsilon}(q_{1},q_{2},p_{1},p_{2})=\tfrac12\left(q_{1}^{2}+p_{1}^{2}\right)+\tfrac{\varepsilon}{2}\left(q_{2}^{2}+p_{2}^{2}\right)+\varepsilon\,U(q_{1},q_{2}),
\qquad
U(q_{1},q_{2})=q_{1}q_{2}\sin(2q_{1}+2q_{2}),
\]
Here, \((q_{1},p_{1})\) forms a unit-frequency fast oscillator, \((q_{2},p_{2})\) forms a slow oscillator with frequency of order \(\varepsilon\), and \(\varepsilon U\) couples the two. Hamilton's equations are
\[
\dot{q}_{1}=p_{1},\qquad
\dot{q}_{2}=\varepsilon p_{2},\qquad
\dot{p}_{1}=-q_{1}-\varepsilon\,\partial_{q_{1}}U,\qquad
\dot{p}_{2}=-\varepsilon q_{2}-\varepsilon\,\partial_{q_{2}}U.
\]
We set \(\varepsilon=10^{-2}\) and integrate \(300\) trajectories for \(3000\) steps using classical RK4 with \(\Delta t=0.1\); this step size resolves the fast period with approximately \(63\) samples. Following Duruisseaux et al.~\cite{duruisseaux2023approximation}, we draw equal numbers of initial conditions from three phase-space windows. The general window uses \(q_{1},p_{1}\sim\mathcal{U}[-1.6,1.6]\), \(q_{2}\sim\mathcal{U}[-0.9,1.6]\), and \(p_{2}\sim\mathcal{U}[-1.1,1.1]\). The second window contains small slow-oscillator orbits centered to the right, with \(q_{2}\sim\mathcal{U}[0.6,1.5]\) and \(p_{2}\sim\mathcal{U}[-0.5,0.5]\); the third contains small orbits around the origin, with \(q_{2},p_{2}\sim\mathcal{U}[-0.8,0.8]\). In both restricted windows, \(q_{1},p_{1}\sim\mathcal{U}[-1.6,1.6]\). We record the total energy \(\mathcal{H}_{\varepsilon}\) at every saved frame.

\subsection{2D Shallow Water Equations (SWE) with Random Initial Condition.}\label{app:swe}
We generate a two-dimensional shallow-water dataset on a periodic square domain with physical parameters
$L_x=L_y=10^6$, gravity $g=9.81$, and mean depth $H=100$.
The computational grid is uniform with $N_x=N_y=64$, and periodicity is enforced via circular shifts in the spatial discretization.
All optional source terms are disabled (Coriolis, $\beta$-effect, friction, wind, sources/sinks), yielding an inviscid, unforced configuration.

The solver evolves conservative variables $(\mathbf{h},\mathbf{m}_{\mathbf{x}},\mathbf{m}_{\mathbf{y}})$ in flux-divergence form, where the mass equation is
\[
\boldsymbol{\partial}_{\mathbf{t}} \mathbf{h} = -\left(\boldsymbol{\partial}_{\mathbf{x}} \mathbf{m}_{\mathbf{x}} + \boldsymbol{\partial}_{\mathbf{y}} \mathbf{m}_{\mathbf{y}}\right),
\]
and the momentum fluxes include the hydrostatic pressure contribution $\tfrac{1}{2} g \mathbf{h}^2$.
Time integration uses a CFL-based step size
\[
\Delta \mathbf{t} = 0.1\,\frac{\min(\Delta \mathbf{x},\Delta \mathbf{y})}{\sqrt{gH}},
\]
together with a fourth-order Runge-Kutta (RK4) scheme.
At each saved step, diagnostic velocities are recovered as $\mathbf{u}=\mathbf{m}_{\mathbf{x}}/\mathbf{h}$ and $\mathbf{v}=\mathbf{m}_{\mathbf{y}}/\mathbf{h}$ (using a small depth floor), and the free-surface anomaly is computed as $\boldsymbol{\eta}=\mathbf{h}-H$.

Initial conditions are random free-surface perturbations $\boldsymbol{\eta}_0$ constructed by averaging two Gaussian random fields and then repeatedly applying Gaussian smoothing until the maximum nearest-neighbor jump falls below a prescribed threshold, enforcing spatial regularity.
We generate $1000$ independent rollouts and save the resulting time series (including $\mathbf{u}$, $\mathbf{v}$, and the stored height field $\mathbf{h}$).
In the continuous setting, the inviscid, unforced shallow-water equations conserve mass and admit an energy invariant; in our implementation, the conservative flux form promotes mass conservation (up to numerical error), while energy is not exactly preserved due to the explicit RK4 time stepping.

\subsection{Gaussian Pulse Initial Condition under SWE}\label{app:pulse}
We generate a two-dimensional shallow-water dataset on a periodic square domain using the same solver and physical configuration described above (uniform $64\times 64$ grid on $[-5\times 10^{5},\,5\times 10^{5}]^2$, periodic boundaries enforced via circular shifts in the spatial discretization, inviscid and unforced). The prognostic fields are the layer depth (free-surface height) $\mathbf{h}(\mathbf{x},\mathbf{y},\mathbf{t})$ and the horizontal velocity components $\mathbf{u}(\mathbf{x},\mathbf{y},\mathbf{t})$, $\mathbf{v}(\mathbf{x},\mathbf{y},\mathbf{t})$ (equivalently the conservative variables $(\mathbf{h},\mathbf{m}_{\mathbf{x}},\mathbf{m}_{\mathbf{y}})$ with $\mathbf{m}_{\mathbf{x}}=\mathbf{h}\mathbf{u}$ and $\mathbf{m}_{\mathbf{y}}=\mathbf{h}\mathbf{v}$). We record time series of $(\mathbf{u},\mathbf{v},\mathbf{h})$ together with the corresponding time stamps $\mathbf{t}$.

Unlike random-field initialization, each rollout here is initialized by a deterministic \emph{localized Gaussian pulse} in the free-surface anomaly $\boldsymbol{\eta}_0$. On the discrete grid, we set
\[
\boldsymbol{\eta}_0(i,j)=A\exp\!\left(-\frac{(i-i_c)^2+(j-j_c)^2}{2\sigma^2}\right),
\]
with amplitude $A=0.1$, width $\sigma=2$ grid cells, and center $(i_c,j_c)$ at the domain midpoint by default. The pulse is converted to the initial height field using the same convention as the implementation (e.g., $\mathbf{h}(\cdot,\cdot,\mathbf{0})=H+\boldsymbol{\eta}_0$ for mean depth $H$), after which the shallow-water equations are integrated forward in time.

We generate $500$ independent rollouts (each starting from the same centered pulse unless the center is randomized) and save the resulting arrays:
\[
\mathbf{u}\in\mathbb{R}^{N_{\mathrm{traj}}\times N_t\times N_y\times N_x},\quad
\mathbf{v}\in\mathbb{R}^{N_{\mathrm{traj}}\times N_t\times N_y\times N_x},\quad
\mathbf{h}\in\mathbb{R}^{N_{\mathrm{traj}}\times N_t\times N_y\times N_x},\quad
\mathbf{t}\in\mathbb{R}^{N_{\mathrm{traj}}\times N_t},
\]
together with auxiliary stored diagnostics (e.g., $\mathrm{hm}$) and the constructed pulse $\boldsymbol{\eta}_0$.

In the continuous setting, the inviscid, unforced shallow-water equations conserve total mass and admit an energy invariant; in our discrete implementation, the flux-form evolution promotes near-conservation of mass up to numerical error, while exact energy conservation is not enforced due to explicit time stepping.

\subsection{Korteweg--de Vries Equation.}\label{app:kdv}
We follow the KdV setup of Eidnes and Lye~\cite{eidnes2024pseudo}, which is also used to evaluate ENO~\citep{tanaka2025energy}. The equation is
\[
u_{t}+\eta\left(\tfrac12 u^{2}\right)_{x}-\gamma^{2}u_{xxx}=0,\qquad \eta=6,\quad \gamma=1,
\]
on the periodic domain \([0,P]\) with \(P=20\), discretized on \(M=100\) uniform grid points. In pseudo-Hamiltonian form the equation reads \(u_{t}=\partial_{x}\,\delta\mathcal{H}/\delta u\) with the conserved functional
\[
\mathcal{H}[u]=-\int\left(\tfrac{\eta}{6}u^{3}+\tfrac{\gamma^{2}}{2}u_{x}^{2}\right)dx .
\]
Initial conditions follow their two-soliton family,
\[
u(x,0)=2\sum_{l=1}^{2}c_{l}^{2}\,\mathrm{sech}^{2}\!\Big(c_{l}\big((x+\tfrac{P}{2}-d_{l}P)\bmod P-\tfrac{P}{2}\big)\Big),
\qquad c_{l}\sim\mathcal{U}(\tfrac12,2),\quad d_{l}\sim\mathcal{U}(0,1),
\]
corresponding to two solitons of heights \(2c_{l}^{2}\) centered at \(d_{l}P\). We use a Fourier pseudospectral discretization and integrating-factor RK4, which treats the stiff dispersive term exactly. Saved frames are separated by \(\Delta t=0.01\), with \(20\) solver substeps per saved interval. We generate \(100\) trajectories up to \(t=2\). Across the generated trajectories, the median relative drift in \(\mathcal{H}\) is \(6.3\times10^{-6}\); doubling the spatial resolution changes the solution by a relative \(L_{2}\) error of \(4.9\times10^{-5}\).
%%%%%%%%%%%%%%%%%%%%%%%%%%%%%%%%%%%%%%%%%%%%%%
\begin{table}[b]
    \vspace{-0.15in}
    \centering
    \caption{\textbf{Model parameter count.} Parameter counts of FS-HNN and the baselines are of the same order of magnitude, so all models are compared at comparable capacity. Some baselines are not scaled further because their accuracy degrades as the networks become deeper and wider.}
    \label{tab:modelsize}
    \vskip 0.03in
    \setlength{\tabcolsep}{4pt}
    \begin{tabular}{lccccccc}
        \toprule
         & FS-HNN & FS-HNN & \multirow{2}{*}{HNN} & \multirow{2}{*}{MLP} & \multirow{2}{*}{SympNet} & \multirow{2}{*}{PHNN} & \multirow{2}{*}{FNO} \\
         & (ODE) & (PDE) & & & & & \\
        \midrule
        Params. & $3.21{\times}10^{6}$ & $2.04{\times}10^{6}$ & $2.82{\times}10^{6}$ & $1.32{\times}10^{6}$ & $0.76{\times}10^{6}$ & $4.80{\times}10^{6}$ & $4.22{\times}10^{6}$ \\
        \bottomrule
    \end{tabular}
\end{table}
\section{Additional Experimental Results}
Supplementary results on the benchmarks generated above follow the main-text division into ODE and PDE systems, covering physical-energy error, additional PDE rollouts, the ablations referenced in Section~\ref{subsec:ablation}, and training cost. All models have parameter counts of the same order of magnitude (Table~\ref{tab:modelsize}), so the comparisons are made at comparable capacity.
%%%%%%%%%%%%%%%%%%%%%%%%%%%%%%%%%%%%%%%%%%%%%%
\subsection{ODE Systems}\label{oderesults}
\begin{table}[t]
    \centering
    \caption{\textbf{Physical-energy error for ODE rollouts.} We report the median relative error over the rollout for the pendulum, double pendulum, and fast–slow oscillator, and the final relative error for FPUT. FS-HNN has the lowest energy error on all four systems, so its rollouts stay faithful to the physical energy as well as to the state.}
    \label{tab:ode_energy_stats}
    \vskip 0.03in
    \begin{tabular}{lcccc}
        \toprule
        Model & Pendulum & Double pendulum & FPUT & Fast--slow\\
        \midrule
        FS-HNN & $\mathbf{1.4\times10^{-3}}$ & $\mathbf{2.3\times10^{-1}}$ & $\mathbf{2.3\times10^{-2}}$ & $\mathbf{5.4\times10^{-4}}$ \\
        HNN & $8.7\times10^{-1}$ & $4.5\times10^{0}$ & $9.2\times10^{3}$ & $1.1\times10^{-2}$ \\
        MLP & $9.5\times10^{-1}$ & $7.9\times10^{0}$ & $4.5\times10^{-2}$ & $1.5\times10^{-2}$ \\
        SympNet & -- & -- & $2.6\times10^{3}$ & $5.3\times10^{-3}$ \\
        \bottomrule
    \end{tabular}
    \vspace{-0.15in}
\end{table}
\begin{table}[b]
    \vspace{-0.15in}
    \centering
    \caption{\textbf{Number of nested views.} We report 50-step rollout MSE from one seed on held-out initial conditions, a protocol that differs from that of Table~\ref{tab:ode}. Fixed cutoffs are obtained by repeated halving from the Nyquist frequency, and spectrum-selected cutoffs are evaluated at $K=3$. Parentheses give the ratio of HNN MSE to the MSE of each configuration, with HNN using the architecture of Table~\ref{tab:ode}. Adding views with halved cutoffs improves FPUT but degrades the fast–slow oscillator, while spectrum-selected cutoffs give the lowest error on both systems.}
    \label{tab:ksweep}
    \vskip 0.03in
    \begin{tabular}{lccccc}
        \toprule
        \multirow{2}{*}{Model} & \multirow{2}{*}{$K$} & \multicolumn{2}{c}{FPUT} & \multicolumn{2}{c}{Fast--slow} \\
        \cmidrule(lr){3-4}\cmidrule(lr){5-6}
        & & MSE & Gain & MSE & Gain \\
        \midrule
        \multicolumn{6}{l}{Baseline} \\
        \cmidrule(l){2-6}
        HNN & -- & $2.03\times10^{-2}$ & $1.0\times$ & $9.64\times10^{-5}$ & $1.0\times$ \\
        \midrule
        \multicolumn{6}{l}{Fixed} \\
        \cmidrule(l){2-6}
        FS-HNN & 1 & $3.37\times10^{-3}$ & $6.0\times$ & $2.29\times10^{-5}$ & $4.2\times$ \\
        FS-HNN & 2 & $3.32\times10^{-3}$ & $6.1\times$ & $3.07\times10^{-5}$ & $3.1\times$ \\
        FS-HNN & 3 & $2.58\times10^{-3}$ & $7.9\times$ & $8.44\times10^{-3}$ & $0.01\times$ \\
        FS-HNN & 4 & $1.38\times10^{-3}$ & $14.7\times$ & $1.21\times10^{-2}$ & $0.008\times$ \\
        \midrule
        \multicolumn{6}{l}{Spectrum} \\
        \cmidrule(l){2-6}
        FS-HNN & 3 & $\mathbf{9.58\times10^{-4}}$ & $\mathbf{21.2\times}$ & $\mathbf{2.26\times10^{-5}}$ & $\mathbf{4.3\times}$ \\
        \bottomrule
    \end{tabular}
\end{table}
\begin{table}[t]
    \centering
    \caption{\textbf{Fixed versus spectrum-selected cutoffs at $K=3$.} We compare rollout MSE using fixed normalized cutoffs $1.0/0.5/0.25$ with the spectrum-selected cutoffs from Section~\ref{subsec:method_freq_ham}, listed in the final column. Gain is the ratio of the fixed-cutoff MSE to the spectrum-selected MSE. Spectrum-selected cutoffs never underperform the fixed cutoffs, and the gain is largest on the fast–slow oscillator, whose spectral scales lie far below the fixed cutoffs.}
    \label{tab:cutoffs}
    \vskip 0.03in
    \begin{tabular}{lcccl}
        \toprule
        Benchmark & Fixed & Spectrum & Gain & Cutoffs \\
        \midrule
        Pendulum & $2.76\times10^{-5}$ & $\mathbf{1.67\times10^{-5}}$ & $1.7\times$ & $1.0/0.05/0.005$ \\
        Double Pendulum & $9.58\times10^{-2}$ & $\mathbf{8.46\times10^{-2}}$ & $1.1\times$ & $1.0/0.692/0.164$ \\
        FPUT & $2.58\times10^{-3}$ & $\mathbf{9.58\times10^{-4}}$ & $2.7\times$ & $1.0/0.0122/0.0012$ \\
        Fast--Slow & $8.44\times10^{-3}$ & $\mathbf{2.26\times10^{-5}}$ & $373\times$ & $1.0/0.016/0.0016$ \\
        KdV & $4.74\times10^{-2}$ & $\mathbf{4.62\times10^{-2}}$ & $1.0\times$ & $1.0/0.88/0.41$ \\
        \bottomrule
    \end{tabular}
    \vspace{-0.15in}
\end{table}
Table~\ref{tab:ode_energy_stats} reports the physical-energy error of each model over the 1000-step rollout. The analytic energy of each system is evaluated on the predicted states and compared with the energy of the reference trajectory, which fluctuates slightly because the reference is generated numerically. The table gives the median relative energy error over the rollout for the pendulum, double pendulum, and fast–slow oscillator, and the final relative energy error for FPUT**, where the energy error of HNN and SympNet grows over the rollout rather than fluctuating about a fixed level.**

Tables~\ref{tab:ksweep} and~\ref{tab:cutoffs} report the ablations on the number of nested views and on cutoff placement discussed in Section~\ref{subsec:ablation}, with the cutoff comparison extended to KdV.
%%%%%%%%%%%%%%%%%%%%%%%%%%%%%%%%%%%%%%%%%%%%%%
\subsection{PDE Systems}\label{pderesults}
\begin{figure}[b]
    \centering
    \begin{subfigure}{0.32\linewidth}
        \centering
        \includegraphics[width=\linewidth]{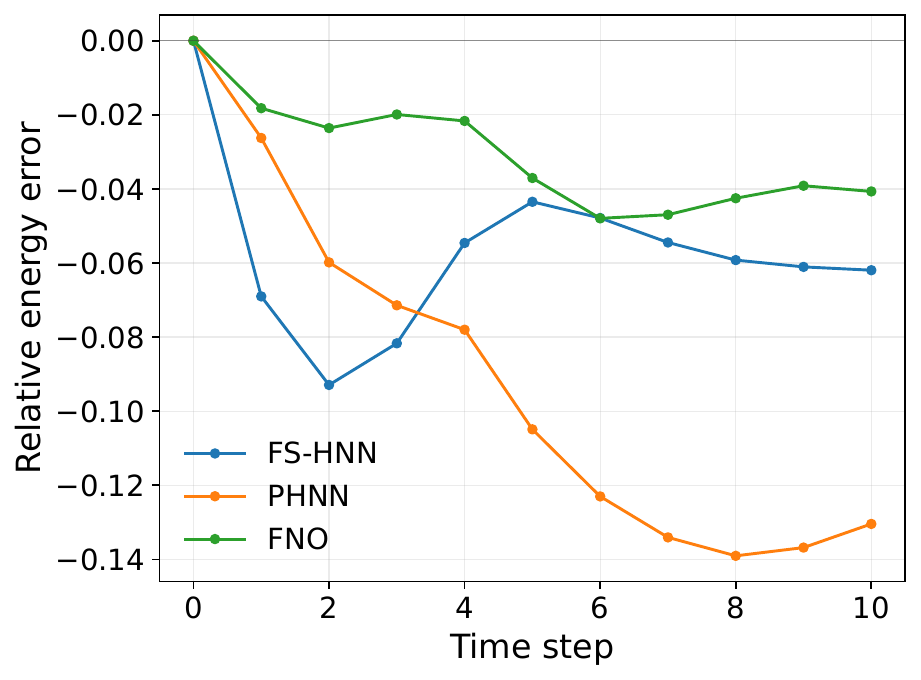}
        \caption{SWE Pulse}
    \end{subfigure}\hfill
    \begin{subfigure}{0.32\linewidth}
        \centering
        \includegraphics[width=\linewidth]{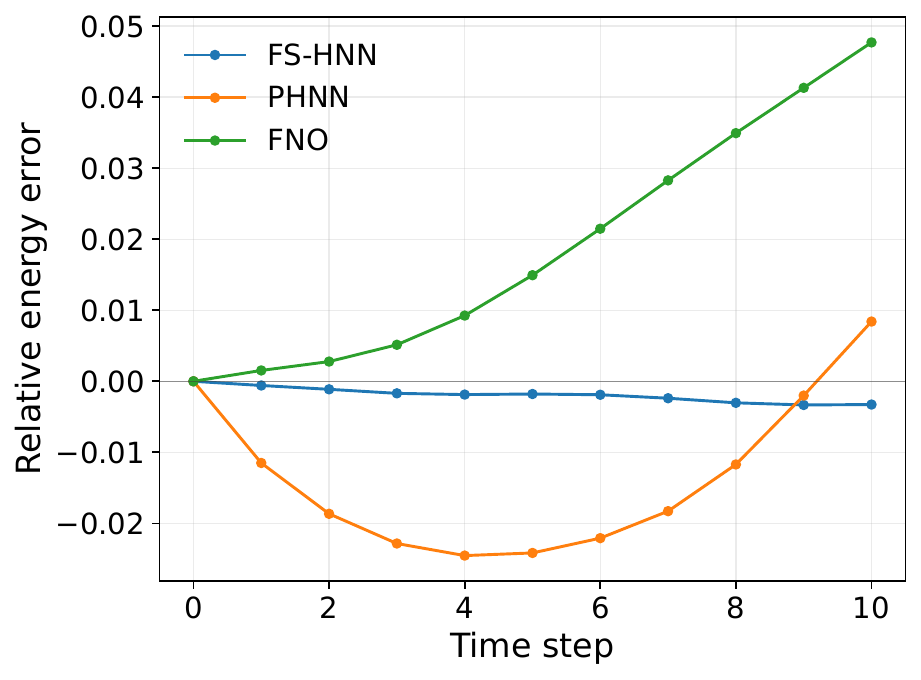}
        \caption{SWE Random}
    \end{subfigure}\hfill
    \begin{subfigure}{0.32\linewidth}
        \centering
        \includegraphics[width=\linewidth]{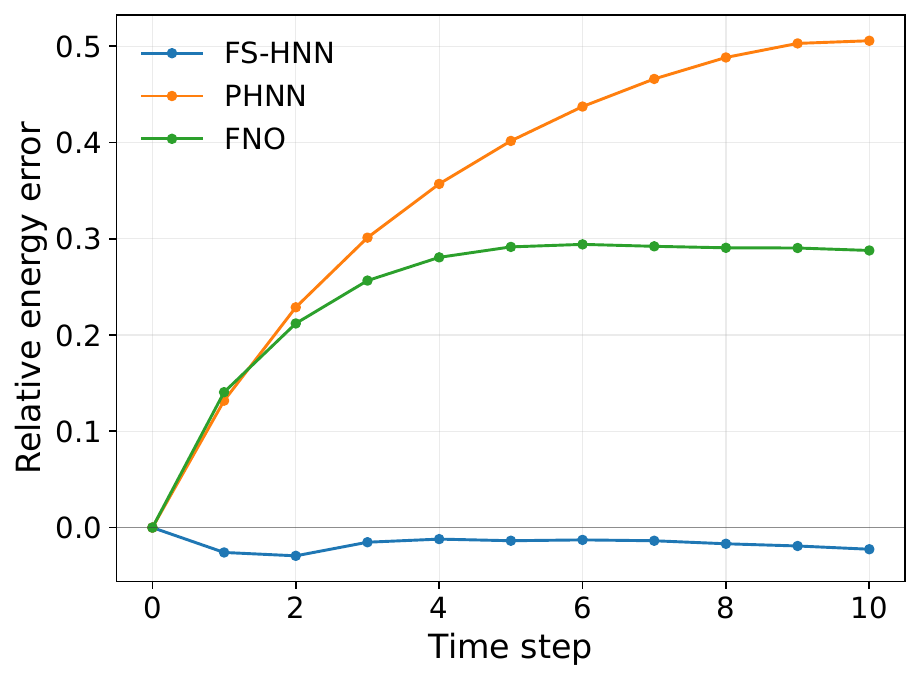}
        \caption{KdV}
    \end{subfigure}
    \caption{\textbf{Physical-energy error for PDE rollouts.} We evaluate the relative error of the SWE energy $\nicefrac{1}{2} h\left(u^{2}+v^{2}\right)+\nicefrac{1}{2} g\eta^{2}$ for (a) Gaussian-pulse and (b) random initial conditions, and of the KdV Hamiltonian $\mathcal{H}[u]$ from Appendix~\ref{app:kdv} for (c) KdV. The reference energies are conserved up to discretization error. Table~\ref{tab:pde_energy_stats} summarizes panels (b) and (c). FS-HNN stays closest to the reference energy on random-initialization SWE and KdV, and ends closer than PHNN on the Gaussian pulse, where FNO tracks the reference most closely.}
    \label{fig:pdeenergy}
\end{figure}
\begin{table}[b]
    \vspace{-0.12in}
    \centering
    \caption{\textbf{Physical-energy error for PDE rollouts.} For random-initialization SWE and KdV, we report relative error at the beginning of the rollout, the median relative error over the rollout, and the final relative error. The reference energy changes by $-8.2\times10^{-3}$ for SWE and $-7.2\times10^{-7}$ for KdV over the respective windows. FS-HNN stays closest to the reference energy on both benchmarks, over the rollout and at its end.}
    \label{tab:pde_energy_stats}
    \vskip 0.03in
    \begin{tabular}{lcccccc}
        \toprule
        & \multicolumn{3}{c}{SWE Random} & \multicolumn{3}{c}{KdV} \\
        \cmidrule(lr){2-4}\cmidrule(lr){5-7}
        Model & $t=0$ & Median & Final & $t=0$ & Median & Final \\
        \midrule
        FS-HNN & $0$ & $\mathbf{1.9\times10^{-3}}$ & $-3.3\times10^{-3}$ & $0$ & $\mathbf{1.6\times10^{-2}}$ & $-2.2\times10^{-2}$ \\
        PHNN & $0$ & $1.8\times10^{-2}$ & $+8.4\times10^{-3}$ & $0$ & $4.2\times10^{-1}$ & $+5.1\times10^{-1}$ \\
        FNO & $0$ & $1.5\times10^{-2}$ & $+4.8\times10^{-2}$ & $0$ & $2.9\times10^{-1}$ & $+2.9\times10^{-1}$ \\
        \bottomrule
    \end{tabular}
\end{table}
Figure~\ref{fig:pdeenergy} shows the physical-energy error of FS-HNN, PHNN, and FNO on all three PDE benchmarks over the first 10 rollout steps. As in the ODE setting, the analytic energy is evaluated on the predicted fields and compared with that of the reference trajectory. Table~\ref{tab:pde_energy_stats} summarizes the energy error for random-initialization SWE and KdV.

Table~\ref{tab:pde_components} reports the ablation from Section~\ref{subsec:ablation} that separates the effects of frequency structure and energy preservation.
%%%%%%%%%%%%%%%%%%%%%%%%%%%%%%%%%%%%%%%%%%%%%%
\subsection{Computational Cost}\label{app:runtime}
All models are trained on an NVIDIA RTX 6000 Pro GPU. Depending on the batch size and epoch configuration, running all benchmarks requires approximately 30 GB of VRAM at minimum, and training typically does not exceed one hour per benchmark. Table~\ref{tab:runtime} reports per-epoch training time on the PDE benchmarks, with FS-HNN reported separately for a single branch and for the combination stage. Although the full model involves multiple training stages, the per-epoch cost of each stage is comparable to that of the baselines on SWE. On KdV, an FS-HNN stage is approximately three times slower than PHNN and five to six times slower than FNO.
%%%%%%%%%%%%%%%%%%%%%%%%%%%%%%%%%%%%%%%%%%%%%%
\section{Impact Statement}\label{app:impact}
This paper focuses on conservative and stable prediction for dynamical systems. We achieve both accuracy and interpretability. Moreover, our results show that, rather than relying solely on high-resolution data, allocating some modeling capacity to coarser-scale information can further improve predictive performance.

The authors do not intend for the methods described harmful behaviors. We do not believe the contribution introduces distinct societal risks beyond those commonly associated with general-purpose ML techniques; nonetheless, we support careful, responsible deployment.
\begin{table}[t]
    \centering
    \caption{\textbf{Frequency structure versus energy preservation.} We report 10-step rollout MSE on random-initialization SWE and KdV. The single-scale conservative HNN retains the DeepONet Hamiltonian, learned $\mathbf{J}_{\theta}$, and orthogonal projection, while the frequency-structured DeepONet instead retains the frequency decomposition and aggregation. Removing either ingredient raises the error on both benchmarks, so FS-HNN relies on frequency structure and energy preservation together.}
    \label{tab:pde_components}
    \vskip 0.03in
    \begin{tabular}{lcc}
        \toprule
        Model & SWE Random & KdV \\
        \midrule
        FS-HNN & $\mathbf{2.38\times10^{-4}}$ & $\mathbf{4.62\times10^{-2}}$ \\
        HNN & $1.55\times10^{-3}$ & $1.47\times10^{-1}$ \\
        DeepONet & $2.62\times10^{-1}$ & $7.77\times10^{-1}$ \\
        \bottomrule
    \end{tabular}
\end{table}
\begin{table}[t]
    \centering
    \caption{\textbf{Per-epoch training time on PDE benchmarks.}
    We report the average training time in seconds per epoch. FS-HNN is reported
    separately for a single branch and for the combination stage.}
    \vskip 0.03in
    \begin{tabular}{lccc}
        \toprule
        Model / Stage & SWE Pulse & SWE Random & KdV \\
        \midrule
        PHNN & $7.83\times10^{-2}$ & $3.51\times10^{-2}$ & $5.08\times10^{-2}$ \\
        FNO  & $6.98\times10^{-2}$ & $5.93\times10^{-2}$ & $2.61\times10^{-2}$ \\
        FS-HNN single branch & $4.09\times10^{-2}$ & $5.56\times10^{-2}$ & $1.31\times10^{-1}$ \\
        FS-HNN combination   & $5.57\times10^{-2}$ & $5.97\times10^{-2}$ & $1.54\times10^{-1}$ \\
        \bottomrule
    \end{tabular}
    \label{tab:runtime}
\end{table}

\phantom{Filling space...}
%%%%%%%%%%%%%%%%%%%%%%%%%%%%%%%%%%%%%%%%%%%%%%
%%%%%%%%%%%%%%%%%%%%%%%%%%%%%%%%%%%%%%%%%%%%%%%
\clearpage
% \input{sections/checklist}
%%%%%%%%%%%%%%%%%%%%%%%%%%%%%%%%%%%%%%%%%%%%%%%
\end{document}